\documentclass{article}

\usepackage{microtype}
\usepackage{graphicx}
\usepackage{booktabs} 
\usepackage[table]{xcolor} 
\usepackage{hyperref}
\usepackage{amsthm}

\usepackage[accepted]{icml2026}
\makeatletter
\def\@notice{}
\makeatother

\usepackage{arydshln}
\usepackage{makecell}
\usepackage{url}
\usepackage{wrapfig}
\usepackage{placeins}
\usepackage{float}
\usepackage{amsfonts}
\usepackage{amssymb}
\usepackage{amsmath}
\usepackage{subcaption}
\usepackage[table]{xcolor}
\definecolor{graybg}{rgb}{0.95,0.95,0.95}
\usepackage{multirow}
\usepackage{threeparttable}
\usepackage{makecell}

\definecolor{highlightblue}{HTML}{E8EFFF}

\icmltitlerunning{RePack then Refine: Efficient Diffusion Transformer with Vision Foundation Model}

\begin{document}

\twocolumn[

\icmltitle{RePack then Refine: Efficient Diffusion Transformer with Vision Foundation Model}
\begin{icmlauthorlist}
\icmlauthor{Guanfang Dong}{huawei}
\icmlauthor{Luke Schultz}{huawei}
\icmlauthor{Negar Hassanpour}{huawei}
\icmlauthor{Chao Gao}{huawei}
\end{icmlauthorlist}

\icmlaffiliation{huawei}{Huawei Technologies Canada Ltd., Edmonton, Alberta, Canada. Emails: \{guanfang.dong, luke.schultz\}@h-partners.com, \{negar.hassanpour2, chao.gao4\}@huawei.com}
\icmlcorrespondingauthor{Guanfang Dong}{guanfang.dong@h-partners.com} 
\icmlkeywords{Diffusion Models, Representation Learning}

\vskip 0.3in
]

\printAffiliationsAndNotice{}

\begin{abstract}
Semantic-rich features from Vision Foundation Models (VFMs) have been leveraged to enhance Latent Diffusion Models (LDMs). However, raw VFM features are typically high-dimensional and redundant, increasing the difficulty of learning and reducing training efficiency for Diffusion Transformers (DiTs). In this paper, we propose Repack then Refine, a three-stage framework that brings the semantic-rich VFM features to DiT while further accelerating learning efficiency. Specifically, the RePack module projects the high-dimensional features onto a compact, low-dimensional manifold. This filters out the redundancy while preserving essential structural information. A standard DiT is then trained for generative modeling on this highly compressed latent space. Finally, to restore the high-frequency details lost due to the compression in RePack, we propose a Latent-Guided Refiner, which is trained lastly for enhancing the image details. On ImageNet-1K, RePack-DiT-XL/1 achieves an FID of 1.82 in only 64 training epochs. With the Refiner module, performance further improves to an FID of 1.65, significantly surpassing latest LDMs in terms of convergence efficiency. Our results demonstrate that packing VFM features, followed by targeted refinement, is a highly effective strategy for balancing generative fidelity with training efficiency. Source code is publicly available at \url{https://github.com/guanfangdong/RePack-then-Refine}.

\vspace{-15pt}
\end{abstract}

\section{Introduction}
\label{sec:Introduction}
\begin{figure}[t]
    \centering
    \includegraphics[width=0.9\linewidth]{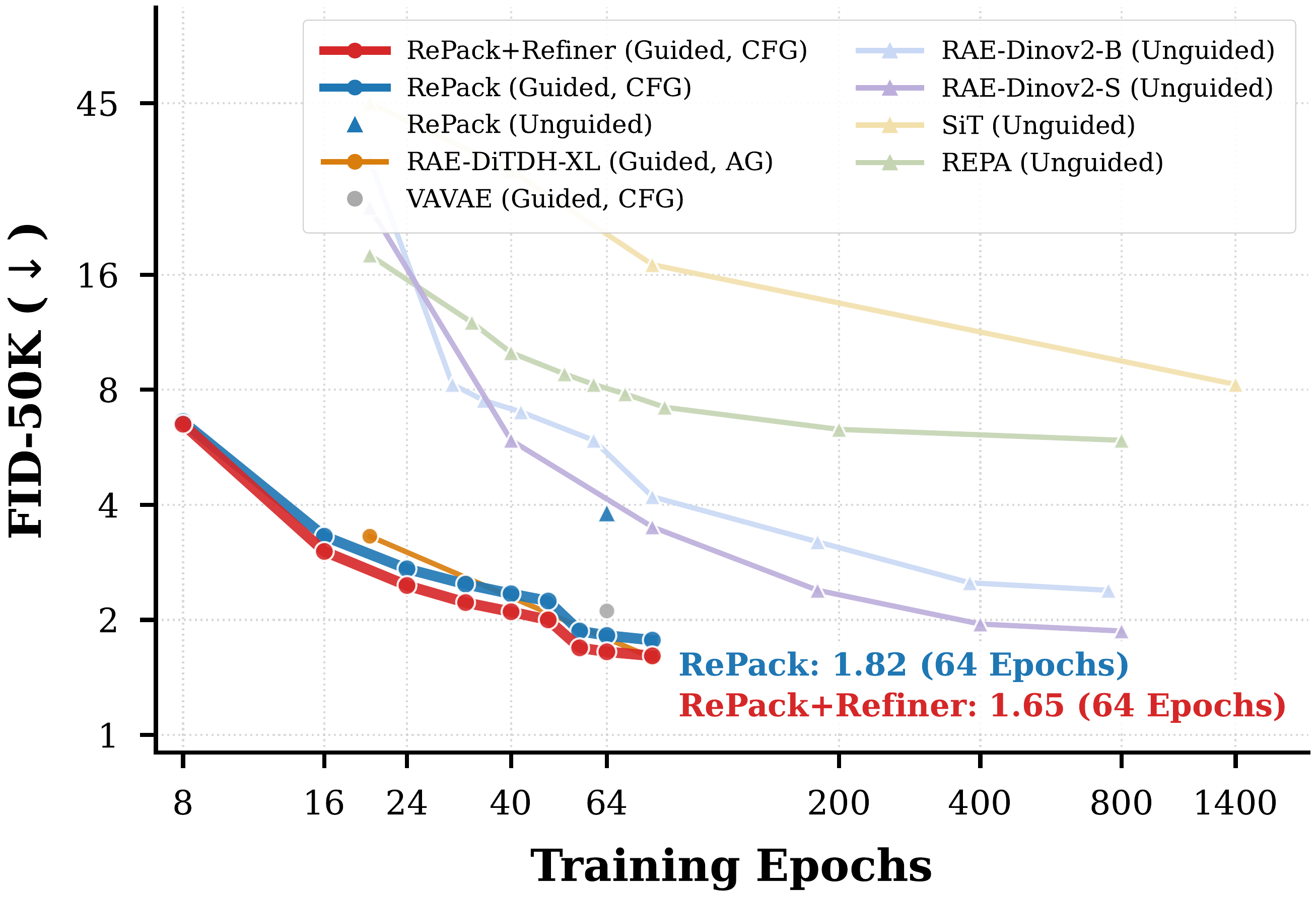}
    \caption{\textbf{FID-50K vs. Training Epochs on ImageNet $256 \times 256$.} RePack achieves strong image generation quality with fewer training epochs. For a fair and clear comparison, models evaluated with guidance (Classifier-Free Guidance / AutoGuidance) are denoted by \textbf{circles}, while unguided models are denoted by \textbf{triangles}. Notably, RePack exhibits a clear drop in FID at 42 epochs, which is attributed to a scheduled learning rate decay. We find that stage-wise learning rate decay leads to lower FID than using a standard learning rate scheduler. Furthermore, we highly recommend using guidance to achieve optimal generation performance.}
    \label{fig:fitting_speed}
    \vspace{-10pt}
\end{figure}
The landscape of Latent Diffusion Models~(LDMs) has evolved significantly since the success of Stable Diffusion~\cite{LDM_22}.
While early models relied on Variational Autoencoders~(VAEs)~\cite{kingma2013auto} to compress images into a latent space, recent methods~\cite{VAVAE, RAE, SVG} directly leverage powerful Vision Foundation Models (VFMs), such as CLIP~\cite{CLIP} and DINOv2/v3~\cite{DinoV2,DinoV3}, as feature encoders.

Algorithms for integrating VFMs generally fall into two categories.
The first category, represented by VA-VAE~\cite{VAVAE}, maintains a standard VAE architecture while using projection heads to align the generative latent space with the VFM feature space during training.
Despite benefiting from strong reconstruction capabilities, these methods often use complicated loss functions for balancing the optimization directions, compromising the rich semantic features of the VFMs.
This ultimately slows down the learning of the Diffusion Transformer (DiT).
The second category, including RAE~\cite{RAE} and SVG~\cite{SVG}, adopts frozen VFM features directly as latent representations under the assumption that richer representations should yield superior generation performance.
Since most encoder parameters remain frozen, these methods are typically easier to train.
Consequently, the practice of feeding these raw, high-density features directly into the Diffusion Transformer~(DiT)~\cite{DiT} has become a new trend. We provide a comprehensive discussion of related work in Appendix~\ref{sec:related_works}.

\begin{figure}[t]
    \centering
    \includegraphics[width=\linewidth]{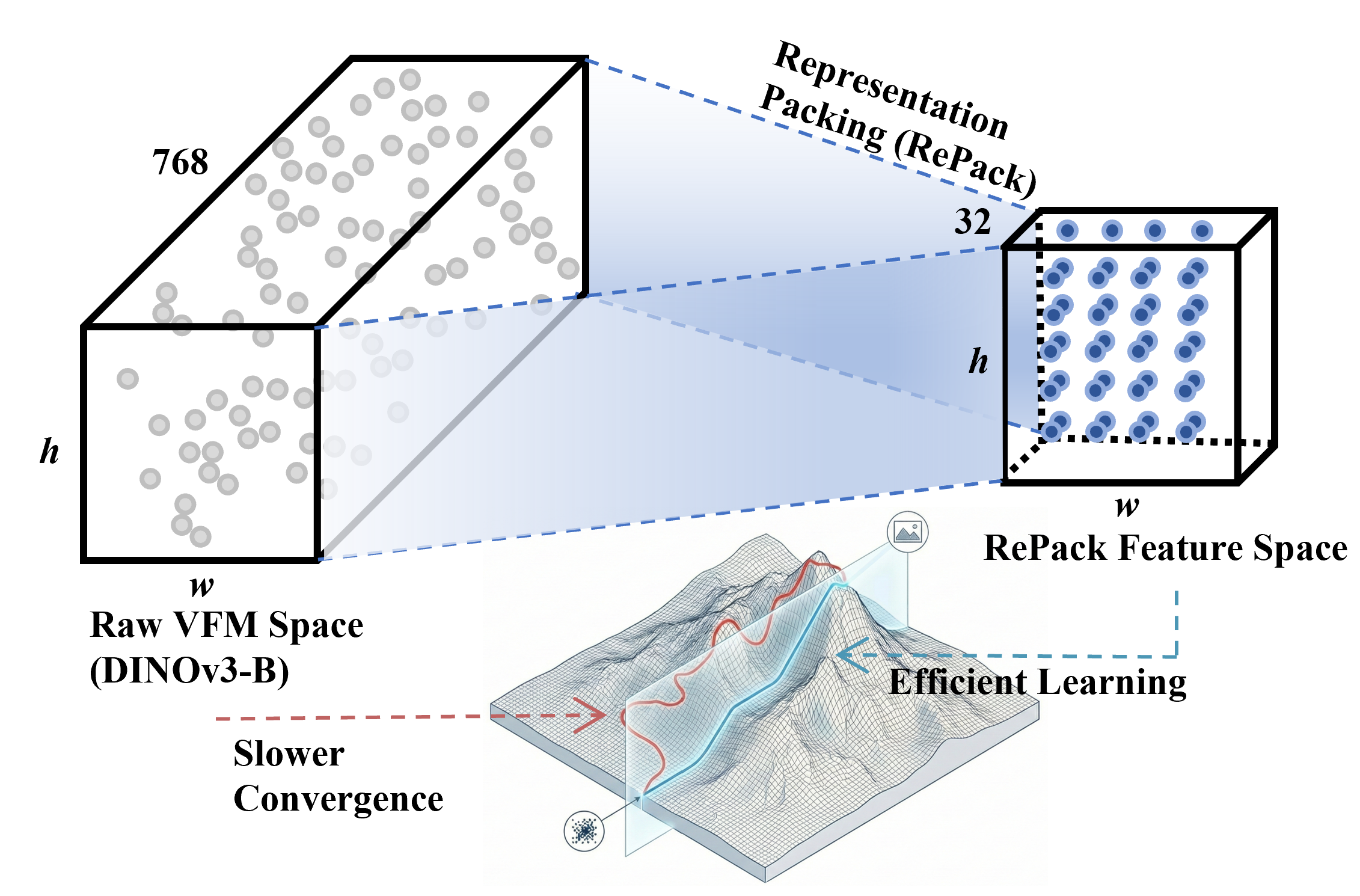}
    \caption{\textbf{Design Philosophy of RePack.} We project high-dimensional raw VFM features (768 channels) into a compact RePack feature space (32 channels). The bottom schematic illustrates learning difficulties: a redundant high-dimensional space leads to slower convergence (red path), while a lower-dimensional semantic space provides a smoother optimization path that enables efficient learning (blue path).
    }
    \label{fig:design_concept}
    \vspace{-15pt}
\end{figure}

In this paper, we argue that direct use of raw VFM features is suboptimal, as it overlooks the fundamental purpose of the encoder: \textbf{information compression}. 
For example, when using a standard ViT encoder like DINOv3-B/16, the extracted feature map typically has dimensions of $(768, H/16, W/16)$.
We observe that the total number of numerical elements in these features is identical to the original image ($768 / 16^2 = 3$, matching the three RGB channels). 
However, since semantic information intrinsically lies on a low-dimensional manifold \cite{Manifold-Assumption}, performing diffusion in such a high-dimensional space is both inefficient and computationally demanding.
The model has to learn a distribution dominated by redundant information that is less important for semantic structure.
As illustrated in Figure \ref{fig:design_concept} (Left), this significantly increases the complexity of distribution modeling.
This can also be extrapolated by comparing the empirical convergences of RAE and VA-VAE. 
Despite the fact that RAE is designed to be semantically richer, in reality, it learns slower than the VFM-supervised VA-VAE, which operates on a compressed 32-channel space (VA-VAE at 64 epochs: 2.11 FID; RAE at 80 epochs: 4.28 FID), as shown in Figure~\ref{fig:fitting_speed}.

Motivated by the need to address this issue, we propose \textbf{RePack} (Representation Packing) that brings back representation compression to LDMs.
As illustrated in Figure \ref{fig:design_concept} (Middle, Right), RePack is designed as a lightweight, bias-free linear projection layer that packs \emph{high-dimensional VFM features} (DINOv3-B, 768-dimensional) into a compact low-dimensional manifold.
With a small number of parameters (0.04M), RePack encourages the encoder to focus on the most principal semantic components of the pre-trained VFM.
This results in a dense semantic representation, leading to easier generative modeling.
In our experiments, RePack-DiT-XL/1 achieves an FID of \textbf{1.82} after only 64 training epochs, significantly outperforming existing methods at the same stage.

To further enhance the image generation fidelity of RePack, we introduce a \textbf{Latent-Guided Refiner}.
Instead of forcing the RePack encoder to learn textures, which would compromise its semantic efficiency, we offload this task to the Refiner. Specifically, we design the Refiner by referring to the latest advances in Pixel Diffusion Models (PDM) \cite{DeCo, PixelDiT}, which explicitly model high-frequency details through specialized pixel-space decoders. 
Our Refiner leverages the generated latent semantics as guidance to synthesize realistic high-frequency textures that are discarded during the packing process. Consequently, our \emph{RePack then Refine} framework harnesses the complementary strengths of latent and pixel spaces.
We design a three-stage training strategy to optimize the RePack, DiT and Refiner modules separately. After RePack successfully learns how to pack VFM features into a compact representation, DiT exhibits faster learning due to the structural coherence in the compact manifold. In the final stage, the Refiner is trained for high-fidelity textural reconstruction.
In our experiments, with \textit{RePack then Refine}, the FID of RePack-DiT-XL/1 is further improved from 1.82 to \textbf{1.65} at 64 DiT training epochs.
To our knowledge, this represents the fastest-converging result among all recent methods. 

Our contributions are highlighted as follows: 
1) We propose \textbf{RePack}, a semantic compression framework. By identifying redundancy in raw VFM features, RePack employs a compact manifold bottleneck to filter out noise while preserving core semantics.
2) We demonstrate the efficiency of generative modeling in this compact space. RePack-DiT-XL/1 achieves a state-of-the-art FID of \textbf{1.82} in just 64 training epochs.
3) We introduce a \textbf{Latent-Guided Refiner} to compensate for the detail loss caused by the lightweight encoder of RePack. By decoupling structure generation from texture refinement, it further improves the performance to an FID of \textbf{1.65}.

\section{Motivation}
\label{sec:method}
\subsection{VFM Feature as Latent for Diffusion}
Latent Diffusion Models (LDMs) typically decouple the generative process into two stages: perceptual compression and generative modeling. Consider an image $x \in \mathbb{R}^{3 \times H \times W}$, in standard LDMs~\cite{LDM_22}, a VAE encoder compresses $x$ into a low-dimensional latent variable $z \in \mathbb{R}^{c \times h \times w }$, where the channel dimension $c$ typically ranges from 4 to 128.
Recent trends propose replacing the trainable VAE encoder with a frozen Vision Foundation Model (VFM). 
Let $\mathcal{E}_\phi$ denote a frozen VFM encoder. The extracted feature $z_{\text{raw}}$ is given by
\begin{equation}
z_{\text{raw}} = \mathcal{E}_\phi(x) \in \mathbb{R}^{D \times h \times w},
\end{equation}
where $(h, w) = (H/p, W/p)$, $p$ is the patch size, and $D$ is the embedding dimension.
For a standard ViT-B/16 model, $D=768$ and $p=16$.
By calculating the total number of elements in $z_{\text{raw}}$, we observe an exact equivalence:
\begin{equation}
h \times w \times D = \frac{H}{16} \times \frac{W}{16} \times 768 = H \times W \times 3.
\end{equation}
Clearly, the total number of elements in the VFM feature is exactly equal to the number of pixels in the original image.
Notably, for larger backbones (e.g., ViT-L with $D=1024$), the feature space even expands beyond the pixel space.
This implies that, in comparison to VAEs, using a VFM as the encoder produces an uncompressed representation.
Recent studies have shown that VFMs possess strong semantic extraction capabilities, which can aid DiT learning. 
However, generative modeling upon an \emph{uncompressed} representation (w.r.t. the number of image pixels) clearly contradicts the core assumption that LDMs should be built upon a compact representation.

\subsection{Redundancy Analysis via PCA}
\label{sec:redundancy_analysis}
\begin{figure}[t]
    \centering
    \includegraphics[width=0.9\linewidth]{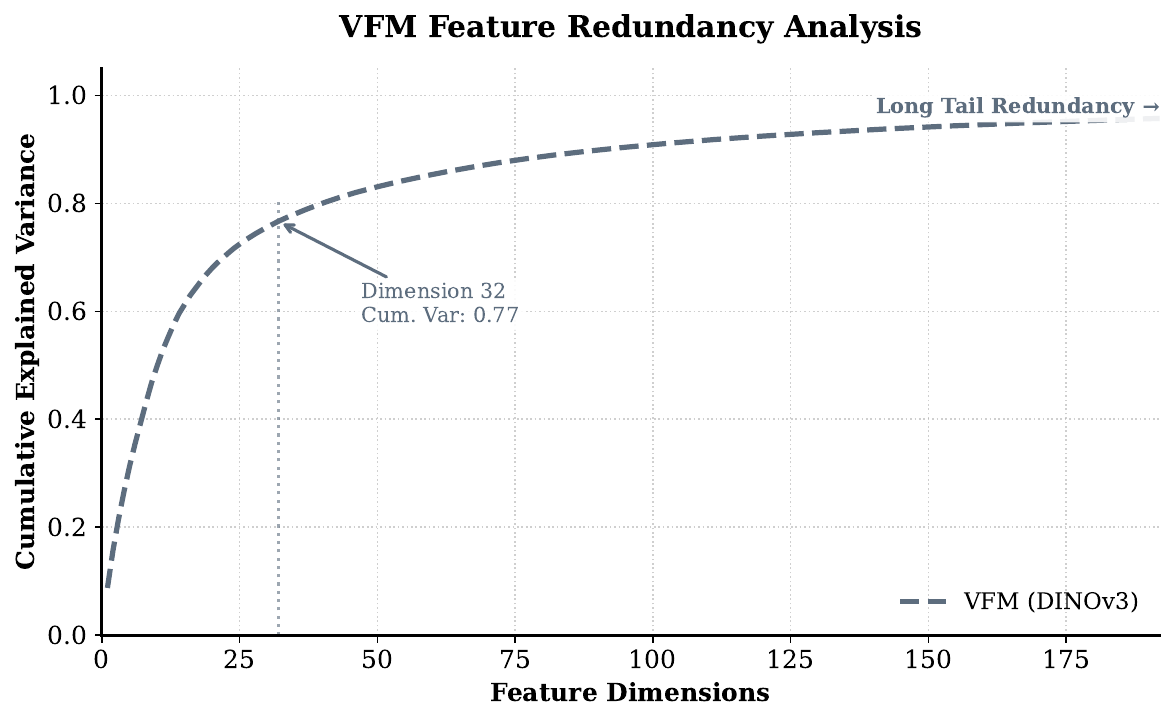}
    \caption{\textbf{Analysis of Representation Redundancy.}
    We perform PCA on features extracted by DINOv3.
    The cumulative explained variance curve reveals a significant long-tail distribution.
    Notably, a distinct \textit{Elbow Point} appears around dimension 32, where the curve begins to flatten.}
    \label{fig:pca_analysis}
\end{figure}

\begin{figure*}[t]
    \centering
    \includegraphics[width=0.96\linewidth]{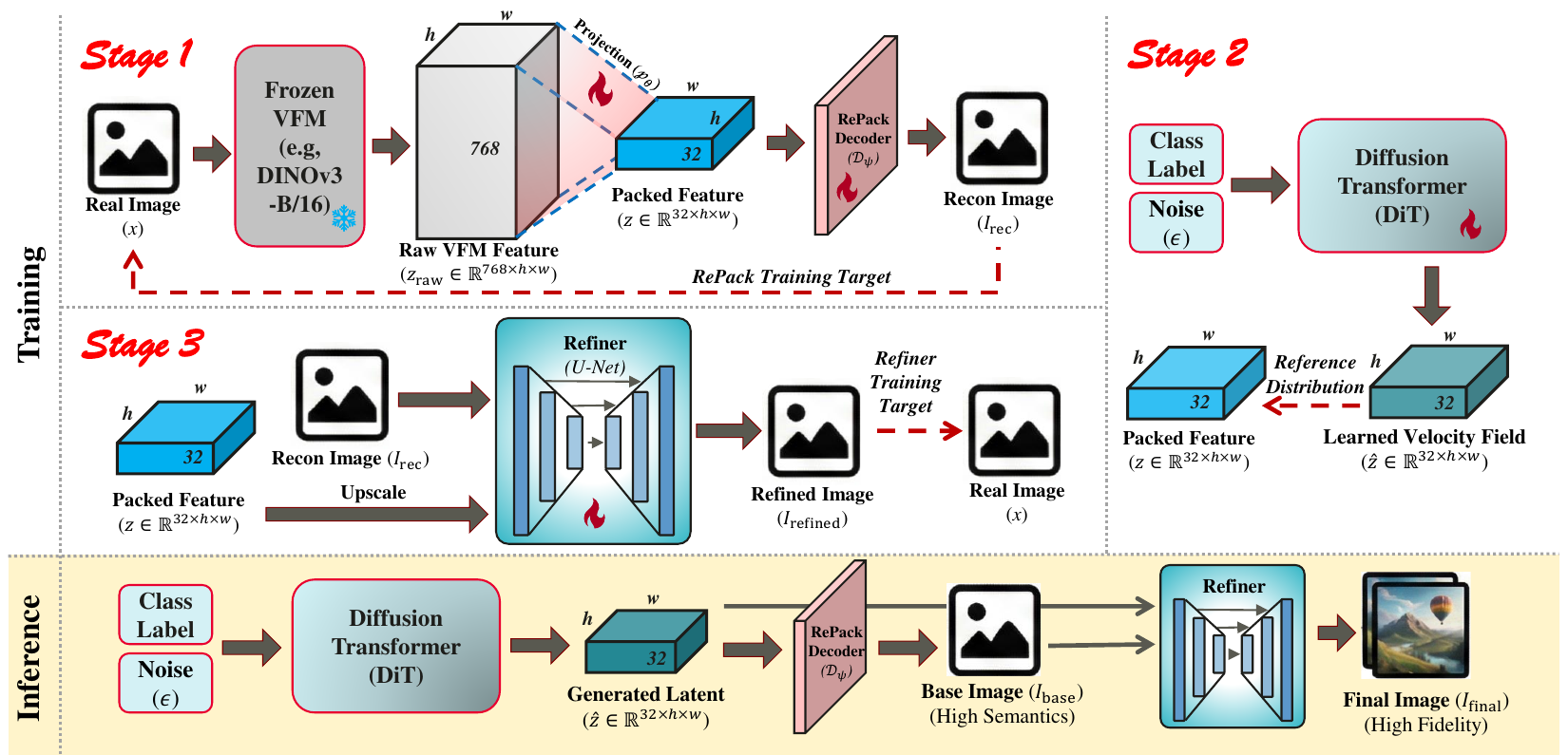}
    \caption{\textbf{The \textit{RePack then Refine} Framework.} The pipeline consists of three stages:
    (1) \textbf{Representation Packing}: Raw VFM features are projected into a compact semantic manifold and reconstructed by the RePack Decoder.
    (2) \textbf{Generative Modeling}: A DiT learns the velocity field over the packed latents.
    (3) \textbf{Latent-Guided Refinement}: A U-Net Refiner restores details using the reconstructed image and upscaled latents.
    The bottom panel shows the inference flow.}
    \label{fig:pipeline_overview}
     \vspace{-10pt}
\end{figure*}

To validate our hypothesis that raw VFM features contain high redundancy due to their high dimensionality, we perform a spectral analysis via Principal Component Analysis (PCA). 
Specifically, we conduct PCA on raw VFM features extracted from the ImageNet-1K validation set using a frozen encoder $\mathcal{E}_\phi$ (i.e., DINOv3-B).
We first extract raw features $z_{\text{raw}} \in \mathbb{R}^{D \times h \times w}$.
We then flatten the spatial dimensions to form a feature matrix $\mathbf{Z}_{\text{raw}} \in \mathbb{R}^{M \times D}$, where $M = h \times w$.
Singular Value Decomposition (SVD) is performed on the centered covariance matrix of $\mathbf{Z}_{\text{raw}}$.
As shown in Figure~\ref{fig:pca_analysis}, the cumulative explained variance exhibits a clear long-tail behavior.
Notably, we observe a distinct elbow in the curve around dimension 32.
The first $\sim32$ principal components account for a substantial majority (approximately 77\%) of the total variance.
Beyond this point, the curve flattens significantly, indicating that the remaining $700+$ dimensions contribute marginally to the variance.
Additional analysis is provided in Appendix~\ref{app:redundancy_analysis}.

This spectral distribution indicates that the effective semantic information in raw VFM features is concentrated in a low-dimensional manifold, while the high-dimensional remainder is dominated by redundancy.
This empirical finding suggests that projecting these features onto a compact subspace may effectively filter out redundancy while preserving core structures.

\section{RePack then Refine framework}
In light of the observation from the redundancy analysis, we propose a \emph{RePack then Refine} framework for more efficient diffusion learning.  

\subsection{Architecture Framework}
As illustrated in Figure~\ref{fig:pipeline_overview}, the \emph{RePack then Refine} framework consists of three independent modules: \textit{Representation Packing}, \textit{Generative Modeling}, and \textit{Latent-Guided Refinement}. 

\textbf{Module for Representation Packing.}
RePack compresses high-dimensional VFM features into a compact manifold through an end-to-end feature mapping and reconstruction pipeline:
\begin{equation}
I_{\text{rec}} = \mathcal{D}_{\psi} \Big( \mathcal{P}_{\theta} \big( \mathcal{E}_{\phi}(x) \big) \Big),
\end{equation}
where $\mathcal{E}_{\phi}$ is a frozen VFM (e.g., DINOv3-B/16), $\mathcal{P}_{\theta}$ denotes a learnable linear projection, and $\mathcal{D}_{\psi}$ is the RePack decoder.
Given an input image $x$, raw features $z_{\text{raw}} \in \mathbb{R}^{D \times h \times w}$ are first extracted by the frozen VFM and then mapped by a lightweight projection layer:
\begin{equation}
z = \mathcal{P}_{\theta}(z_{\text{raw}}), \quad z \in \mathbb{R}^{d \times h \times w},
\end{equation}
where $d \ll D$ (e.g., $d=32$), and $h = H/16, w = W/16$ correspond to the spatial dimensions after patchification.
The projection layer is bias-free, which empirically favors learning a pure subspace projection and leads to improved generation quality.

\textbf{Module for Generative Modeling.}
For the generative backbone, we employ LightningDiT~\cite{VAVAE}. It is an optimized variant of the standard DiT, which improves convergence efficiency by integrating RMSNorm, SwiGLU, rotary positional embeddings (RoPE), and Rectified Flow. It has been widely adopted as a backbone in related frameworks such as VA-VAE and RAE.

\textbf{Module for Latent-Guided Refinement.}
The Refiner adopts a U-Net architecture with skip connections for multi-scale detail propagation. 
Given a latent feature $z \in \mathbb{R}^{32 \times h \times w}$ and its decoded image $I_{\text{rec}} \in \mathbb{R}^{3 \times H \times W}$, we bilinearly upsample $z$ to $z_{\text{up}} \in \mathbb{R}^{32 \times H \times W}$ and concatenate it with $I_{\text{rec}}$ along the channel dimension.
The resulting input $X_{\text{in}} \in \mathbb{R}^{35 \times H \times W}$ is processed by the Refiner to produce the final output:
\begin{equation}
I_{\text{refined}} = \mathcal{R}(X_{\text{in}}) = \mathcal{R}([I_{\text{rec}}, z_{\text{up}}]).
\end{equation}

\subsection{Multi-Stage Training}
\label{sec:training}
We optimize the framework in three independent stages.

\textbf{Stage 1: Representation Packing}

In the first stage, we train the projector $\mathcal{P}_{\theta}$ and decoder $\mathcal{D}_{\psi}$ end-to-end while keeping the VFM $\mathcal{E}_{\phi}$ frozen.
The objective is to maximize the reconstruction quality of $I_{\text{rec}}$ while maintaining a smooth latent manifold.
We employ a composite loss $\mathcal{L}_{\text{RePack}}$:
\begin{equation}
\mathcal{L}_{\text{RePack}} =
\mathcal{L}_{1}
+ \lambda_{\text{lpips}} \mathcal{L}_{\text{lpips}}
+ \lambda_{\text{adv}} \mathcal{L}_{\text{adv}}
+ \lambda_{W} \mathcal{L}_{\text{Watson}}
+ \lambda_{f} \mathcal{L}_{\text{focal}}.
\end{equation}
Here, $\mathcal{L}_{1}$ is the standard pixel-level reconstruction loss.
$\mathcal{L}_{\text{lpips}}$ denotes the LPIPS perceptual loss.
$\mathcal{L}_{\text{adv}}$ represents the adversarial loss from a discriminator to improve perceptual realism.
Additionally, $\mathcal{L}_{\text{Watson}}$~\cite{watson-loss} and $\mathcal{L}_{\text{focal}}$~\cite{focal-freq-loss} are introduced to enhance visual sensitivity and spectral consistency, respectively.

\textbf{Stage 2: Generative Modeling}

In the second stage, we freeze the RePack encoder and decoder and train the LightningDiT backbone.
The generative process is formulated as a flow-matching objective, where the model learns to predict the velocity field that transforms a standard Gaussian distribution into the data distribution.
The training loss minimizes the mean squared error between the predicted and target velocities:
\begin{equation}
\mathcal{L}_{\text{diff}} =
\mathbb{E}_{z, \epsilon, t, c}
\left[
\left\|
\mathbf{v}_t - \mathbf{v}_{\omega}(z_t, t, c)
\right\|_2^2
\right].
\end{equation}
Here, $z$ denotes the clean packed latent features, $\epsilon \sim \mathcal{N}(0, \mathbf{I})$ is Gaussian noise, and $t \in [0,1]$ is sampled from a logit-normal distribution.
The noisy latent is defined as $z_t = t z + (1 - t)\epsilon$, with target velocity $\mathbf{v}_t = z - \epsilon$.
$\mathbf{v}_{\omega}$ denotes the Diffusion Transformer parameterized by $\omega$, which predicts the velocity given $(z_t, t, c)$.

\newcommand{\cc}{\cellcolor{highlightblue}}
\begin{table*}[th!]
\centering
\caption{Class-conditional Image Generation Performance on ImageNet $256 \times 256$.}
\label{tab:repack_full_exp_results}
\small
\setlength{\tabcolsep}{2.5pt}
\begin{tabular}{l|c|c|c@{\hspace{1pt}}@{\hspace{0.5pt}}c@{}@{}c@{\hspace{1pt}}@{\hspace{0.5pt}}c@{}@{}c|ccccc}
\hline
\multirow{2}{*}{Method} & \multirow{2}{*}{\makecell{Training \\ Epochs}} & \multirow{2}{*}{\makecell{Backbone \\ Params (M)}} & \multicolumn{5}{c|}{VAE Params (M)} & \multicolumn{5}{c}{Generation Performance} \\
\cmidrule(lr){4-8} \cmidrule(lr){9-13}
& & & Enc. & Enc$_{tr}$$^\dagger$ & Dec. & Total$_{tr}$$^\dagger$ & Total & gFID$\downarrow$ & sFID$\downarrow$ & IS$\uparrow$ & Pre.$\uparrow$ & Rec.$\uparrow$ \\
\hline
\multicolumn{13}{c}{\textit{Pixel Diffusion Models (PDM)}} \\
\hline
FractalMAR-H \cite{MAR-H} & 600 & 848 & -- & -- & -- & -- & -- & 6.15 & -- & 348.9 & 0.81 & 0.46 \\
PixelFlow \cite{PixelFlow} & 320 & 677 & -- & -- & -- & -- & -- & 1.98 & 5.83 & 282.1 & 0.81 & 0.60 \\
PixNerd \cite{PixelNerd} & 320 & 700 & -- & -- & -- & -- & -- & 1.95 & 4.54 & 300.0 & 0.80 & 0.60 \\
JiT \cite{JiT} & 600 & 953 & -- & -- & -- & -- & -- & 1.86 & -- & 303.0 & -- & -- \\
\multirow{2}{*}{DeCo \cite{DeCo}}& 320 & 682 & -- & -- & -- & -- & -- & 1.90 & 4.47 & 303.0 & 0.80 & 0.61 \\
& 600 & 682 & -- & -- & -- & -- & -- & 1.69 & 4.59 & 304.0 & 0.79 & 0.63 \\
\multirow{2}{*}{PixelDiT \cite{PixelDiT}} & 80 & 797 & -- & -- & -- & -- & -- & 2.36 & 5.11 & 282.3 & 0.80 & 0.57 \\
& 320 & 797 & -- & -- & -- & -- & -- & 1.61 & 4.68 & 292.7 & 0.78 & 0.64 \\
\hline
\multicolumn{13}{c}{\textit{AutoRegressive (AR) Models}} \\
\hline
LlamaGen \cite{LlamaGen} & 300 & 3100 & -- & -- & -- & -- & 72 & 2.18 & 5.97 & 263.3 & 0.81 & 0.58 \\
VAR \cite{VAR} & 350 & 2000 & -- & -- & -- & -- & -- & 1.80 & -- & 365.4 & 0.83 & 0.57 \\
MagViT-v2 \cite{MagViT} & 1080 & 307 & -- & -- & -- & -- & -- & 1.78 & -- & 319.4 & -- & -- \\
MAR \cite{MAR} & 800 & 945 & -- & -- & -- & -- & -- & 1.55 & -- & 303.7 & 0.81 & 0.62 \\
\hline
\multicolumn{13}{c}{\textit{Latent Diffusion Models (LDM)}} \\
\hline
MaskDiT \cite{MaskDiT} & 1600 & 675 & 34 & 34 & 50 & 84 & 84 & 2.28 & 5.67 & 276.6 & 0.80 & 0.61 \\
DiT \cite{DiT} & 1400 & 675 & 34 & 34 & 50 & 84 & 84 & 2.27 & 4.60 & 278.2 & \textbf{0.83} & 0.57 \\
\multirow{2}{*}{VA-VAE \cite{VAVAE}} & 64 & 675 & 28 & 28 & 41 & 70 & 70 & 2.11 & 4.16 & 252.3 & 0.81 & 0.58 \\
& 800 & 675 & 28 & 28 & 41 & 70 & 70 & 1.35 & \textbf{4.15} & 295.3 & 0.79 & 0.65 \\
SiT \cite{SiT} & 1400 & 675 & 34 & 34 & 50 & 84 & 84 & 2.06 & 4.50 & 270.3 & 0.82 & 0.59 \\
FasterDiT \cite{FasterDiT} & 400 & 675 & 34 & 34 & 50 & 84 & 84 & 2.03 & 4.63 & 264.0 & 0.81 & 0.60 \\
\multirow{3}{*}{FAE \cite{FAE}} & 64 & 675 & 1138 & 38 & 476 & 514 & 1614 & 2.01 & 4.39 & 250.3 & 0.83 & 0.59 \\
& 80 & 675 & 1138 & 38 & 476 & 514 & 1614 & 1.92 & 4.35 & 249.6 & 0.83 & 0.59 \\
& 800 & 675 & 1138 & 38 & 476 & 514 & 1614 & 1.41 & 4.34 & 274.1 & 0.81 & 0.63 \\
\multirow{3}{*}{FAE w/ Time. Shift \cite{FAE}} & 64 & 675 & 1138 & 38 & 476 & 514 & 1614 & 1.87 & 4.39 & 241.1 & 0.82 & 0.59
\\& 80 & 675 & 1138 & 38 & 476 & 514 & 1614 & 1.70 & 4.33 & 243.8 & 0.82 & 0.61 \\
& 800 & 675 & 1138 & 38 & 476 & 514 & 1614 & 1.29 & 4.32 & 268.0 & 0.80 & 0.64 \\
MDT \cite{MDT} & 1300 & 675 & 34 & 34 & 50 & 84 & 84 & 1.79 & 4.57 & 283.0 & 0.81 & 0.61 \\
MDTv2 \cite{MDT} & 1080 & 675 & 34 & 34 & 50 & 84 & 84 & 1.58 & 4.52 & \textbf{314.7} & 0.79 & 0.65 \\
REPA \cite{REPA} & 800 & 675 & 34 & 34 & 50 & 84 & 84 & 1.42 & 4.70 & 305.7 & 0.80 & 0.65 \\
RAE (DiT-XL) \cite{RAE} & 800 & 676 & 87 & 0 & 415 & 415 & 502 & 1.41 & -- & 309.4 & 0.80 & 0.63 \\
RAE (DiTDH-XL) \cite{RAE} & 800 & 839 & 87 & 0 & 415 & 415 & 502 & \textbf{1.13} & -- & 262.6 & 0.78 & \textbf{0.67} \\
\multirow{2}{*}{SVG-XL$^\ast$ \cite{SVG}} & 500 & 675 & 40 & 11 & 43 & 54 & 83 & 2.10 & -- & 258.7 & -- & -- \\
& 1400 & 675 & 40 & 11 & 43 & 54 & 83 & 1.92 & -- & 264.9 & -- & -- \\
\hline
\cc & \cc\textbf{64} & \cc & \cc & \cc & \cc & \cc & \cc & \cc\textbf{1.82} & \cc\textbf{4.41} & \cc 269.6 & \cc\textbf{0.83} & \cc\textbf{0.59} \\
\cc\multirow{-2}{*}{\textbf{RePack}} & \cc\textbf{80} & \cc\multirow{-2}{*}{\textbf{675}} & \cc\multirow{-2}{*}{\textbf{85.71}} & \cc\multirow{-2}{*}{\textbf{0.04}} & \cc\multirow{-2}{*}{\textbf{41.42}} & \cc\multirow{-2}{*}{\textbf{41.46}} & \cc\multirow{-2}{*}{\textbf{127.13}} & \cc\textbf{1.77} & \cc\textbf{4.42} & \cc\textbf{273.8} & \cc\textbf{0.82} & \cc\textbf{0.59} \\
\cc & \cc\textbf{64} & \cc & \cc & \cc & \cc & \cc & \cc & \cc\textbf{1.65} & \cc\textbf{4.35} & \cc 268.9 & \cc\textbf{0.81} & \cc\textbf{0.59} \\
\cc\multirow{-2}{*}{\textbf{RePack + Refiner}} & \cc\textbf{80} & \cc\multirow{-2}{*}{\textbf{675}} & \cc\multirow{-2}{*}{\textbf{85.71}} & \cc\multirow{-2}{*}{\textbf{0.04}} & \cc\multirow{-2}{*}{\textbf{41.42}} & \cc\multirow{-2}{*}{\textbf{97.04}} & \cc\multirow{-2}{*}{\textbf{182.71}} & \cc\textbf{1.61} & \cc\textbf{4.40} & \cc\textbf{272.6} & \cc\textbf{0.81} & \cc\textbf{0.60} \\
\hline
\end{tabular}
\begin{tablenotes}
    \footnotesize
    \item $*$ SVG uses 25-step Euler sampling. 
    \item $\dagger$ Enc$_{tr}$ and Total$_{tr}$ denote the number of trainable parameters in the encoder and the overall VAE pipeline, respectively.
\end{tablenotes}
\vspace{-10pt}
\end{table*}

\textbf{Stage 3: Latent-Guided Refinement}

In the final stage, we train the Refiner $\mathcal{R}$ in an \textbf{offline} manner, decoupled from the DiT.
We construct training pairs using ground-truth images $x$.
First, we extract the latent $z = \mathcal{P}_{\theta}(\mathcal{E}_{\phi}(x))$ and reconstruct the base image $I_{\text{rec}} = \mathcal{D}_{\psi}(z)$.
During training, the concatenated input $(I_{\text{rec}}, z_{\text{up}})$ is fed into the Refiner.
The model is optimized to recover the high-frequency details of $x$ using a composite objective:
\begin{equation}
\mathcal{L}_{\text{refine}} =
\mathcal{L}_{1}
+ \lambda_{\text{lpips}} \mathcal{L}_{\text{lpips}}
+ \lambda_{\text{adv}} \mathcal{L}_{\text{adv}}.
\end{equation}
Here, $\mathcal{L}_{\text{adv}}$ denotes the adversarial loss computed using a PatchGAN~\cite{Patch-GAN} discriminator, which focuses on improving local texture realism at the patch level.
By training on reconstruction pairs, the Refiner learns a robust mapping from coarse semantics to fine details, which generalizes well to DiT-generated latents during inference.
\begin{figure*}[t]
    \centering
    \includegraphics[width=0.85\linewidth]{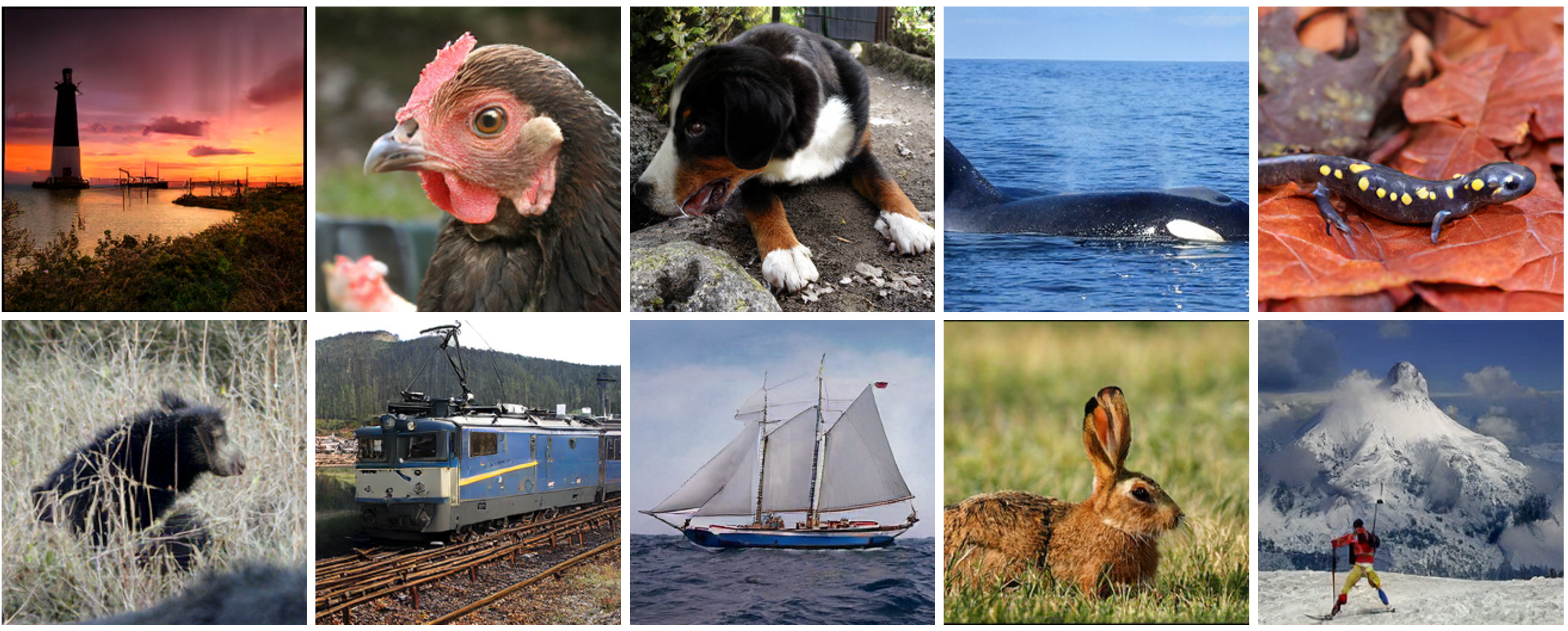}
    \caption{\textbf{Qualitative Results on ImageNet $256 \times 256$ after 64 Training Epochs.} 
    Samples generated by RePack-DiT-XL/1 equipped with the Latent-Guided Refiner. }
    \label{fig:generated_samples_64ep}
    \vspace{-10pt}
\end{figure*}

\subsection{Inference Pipeline}
\label{sec:inference}
As shown in the bottom panel of Figure~\ref{fig:pipeline_overview}, image generation is performed using the three trained modules. We first apply the Diffusion Transformer (DiT). Given a class condition $c$ and Gaussian noise $\epsilon \sim \mathcal{N}(0, \mathbf{I})$, it generates a latent representation $\hat{z} \in \mathbb{R}^{32 \times h \times w}$ by integrating the learned velocity field, capturing the global semantic structure of the target image.
The generated latent $\hat{z}$ is then decoded into pixel space by the RePack decoder $\mathcal{D}_{\psi}$, producing a base image $I_{\text{base}} = \mathcal{D}_{\psi}(\hat{z})$. Due to the compact latent representation, $I_{\text{base}}$ preserves the overall semantic layout.
Finally, the Refiner takes $I_{\text{base}}$ together with the upsampled latent $\hat{z}_{\text{up}}$ as input and restores high-frequency details conditioned on this structural guidance. The resulting image $I_{\text{final}}$ exhibits improved visual fidelity with more realistic textures.

\section{Experimental Results}
\subsection{Implementation Details}
We evaluate our framework on ImageNet-1K \cite{image-net} at $256 \times 256$ resolution. 
We utilize a frozen DINOv3 ViT-B/16~\cite{DinoV3} as the feature extractor, followed by the projector and a standard VA-VAE decoder~\cite{VAVAE}. 
RePack and the Refiner are trained for 20 epochs. 
For generative modeling, we employ DiT-XL/1 and adopt Classifier-Free Guidance (CFG) during inference by default. 
Comprehensive details regarding model architecture, loss configurations, and additional experimental settings are provided in Appendix \ref{sec:arch_details} and \ref{ref:detail_config}.

\subsection{Efficiency and Quality Comparison on Image Generation}
\textbf{Overall Results.} Table \ref{tab:repack_full_exp_results} presents the class-conditional generation performance of RePack on ImageNet $256 \times 256$.   Remarkably, RePack achieves the fastest learning among all algorithms.  
With only 64 epochs of training, RePack-DiT-XL/1 achieves an FID of 1.82. 
By contrast, conventional LDMs (e.g., DiT-XL \cite{DiT}) require 1{,}400 epochs to reach an FID of 2.27. 
This shows that a compact and clean latent space can substantially improve training efficiency.  
Table~\ref{tab:repack_full_exp_results} also shows that, by adding the Latent-Guided Refiner, results are further enhanced: FID is reduced from 1.82 to 1.65, sFID drops to 4.35 and IS reaches 268.9.  This confirms the effectiveness of the Refiner.
Figure~\ref{fig:generated_samples_64ep} contains a few sample images from \textbf{RePack + Refiner}, demonstrating fine-grained details. 

\subsubsection{Comparison with Other VFM Adaptation Methods}
Unlike FAE~\cite{FAE}, which relies on a computationally expensive encoder (1,138M parameters), or PS-VAE~\cite{PS-VAE} which unfreezes the VFM, RePack prioritizes efficiency and practical utility.
FAE’s large parameter count leads to significant memory and latency overheads.
Similarly, fine-tuning the VFM (as in PS-VAE) risks catastrophic forgetting, degenerating a robust foundation model into a dataset-specific encoder.
By contrast, RePack uses a lightweight non-destructive projector. This minimalist design avoids the heavy computational burden of complex encoders while preserving the VFM's universal feature space, enabling the DiT to converge significantly faster.

\subsubsection{Comparison with Raw Feature and Feature Alignment Methods}
Raw-feature methods such as RAE~\cite{RAE} and SVG~\cite{SVG} converge slowly due to high-dimensional sparse representations. 
RePack is significantly more efficient. For a unguided settings, RePack achieves an FID of 3.81 after 64 epochs with only 41M learnable parameters, outperforming RAE-DiT-XL (4.28 at 80 epochs, 415M). When standard Classifier-Free Guidance is applied, RePack efficiently reaches an FID of 1.82 in 64 epochs, whereas SVG requires 500 epochs to reach 2.10 (54M).
VA-VAE~\cite{VAVAE} relies on explicit feature alignment, requiring balanced semantic alignment with pixel reconstruction, and reaches an FID of 2.11 after 64 epochs. RePack instead uses a linear projection that preserves the semantic manifold, yielding a cleaner representation while bypassing optimization conflicts.

\subsubsection{Comparison with Pixel Diffusion Models}
As shown in Table~\ref{tab:repack_full_exp_results}, our \textbf{RePack + Refine} approach achieves stronger results than all six pixel diffusion models (PDMs) on primary generation quality metrics (FID and sFID). For instance, DeCo~\cite{DeCo} requires 600 epochs to achieve an FID of 1.69.
Combining the efficiency of a compact latent space with latent-targeted refinement, our algorithm achieves an FID of 1.65 in 64 epochs. The slow learning of PDMs is due in large part to their operation upon full image pixel space. 
Designed by borrowing the core strength of PDMs, our latent-guided Refiner decouples high-frequency detail modeling from semantic generation, improving visual detail while retaining the efficiency of latent diffusion models.

\subsection{Latent-Guided Refiner Enhances the Modeling of High-Frequency Details}

As discussed in Section~\ref{sec:method}, RePack prioritizes semantic purity and training efficiency by projecting features onto a compact semantic manifold.
While this design significantly accelerates DiT convergence, achieving a gFID of 1.82 within 64 training epochs, the extremely low-dimensional bottleneck ($d=32$) inevitably acts as a low-pass filter, leading to the loss of high-frequency texture information.
This effect is reflected in the reconstruction FID (rFID) of 0.83 for the base RePack model.

\begin{figure}[t]
    \centering
    \includegraphics[width=1\linewidth]{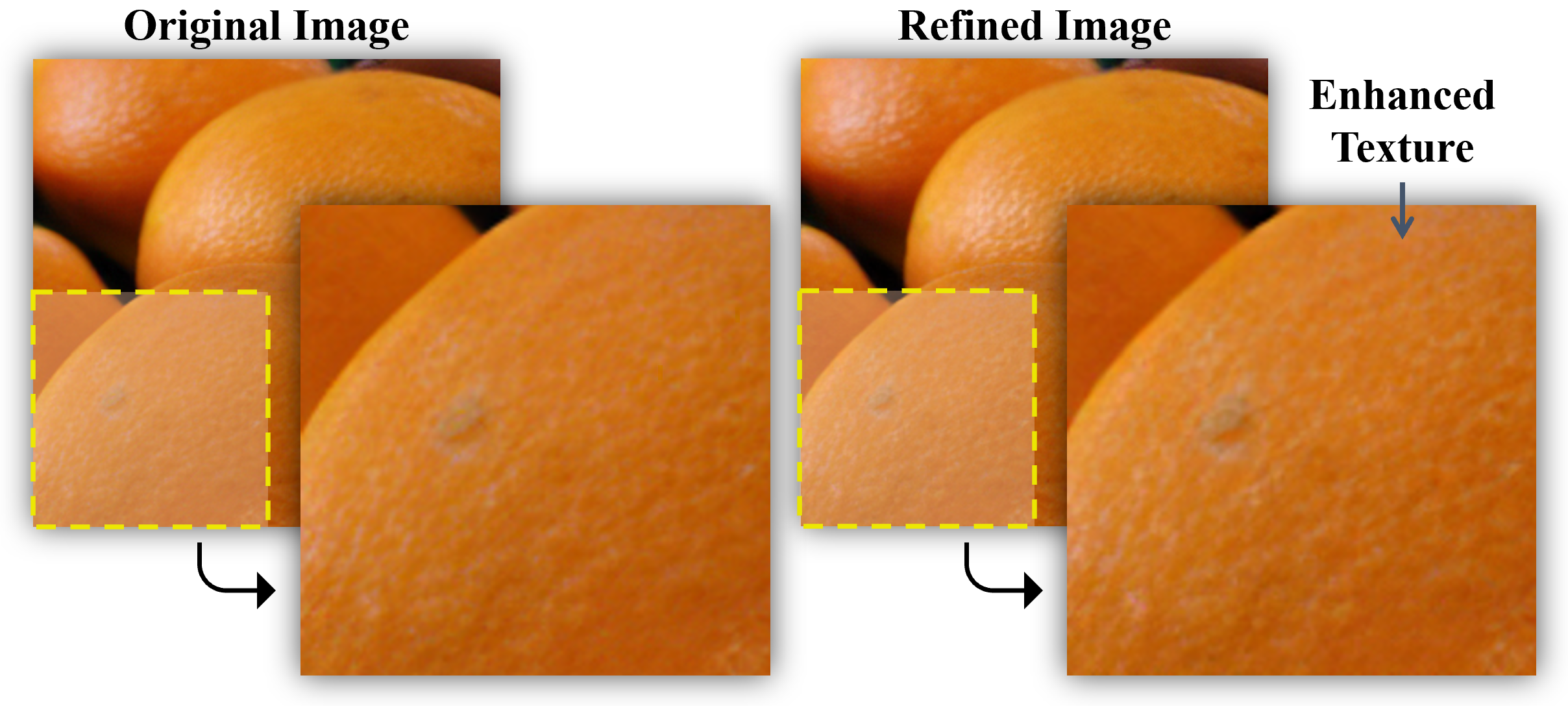}
    \caption{\textbf{Effect of Latent-Guided Refinement.} 
    Visual comparison between the base image decoded from RePack latents (Left) and the refined output (Right).
    As highlighted in the zoomed-in regions, the Refiner recovers realistic high-frequency details, such as the fine porous texture of the orange peel.}
    \label{fig:refiner_qualitative}
    \vspace{-10pt}
\end{figure}

The Latent-Guided Refiner is specifically designed to address this trade-off.
By leveraging the generated latents as structural guidance and explicitly modeling high-frequency details in the pixel domain, the Refiner effectively restores the missing textures.
Enabling the Refiner reduces the rFID from 0.83 to 0.69.
Correspondingly, the generative performance (gFID) of RePack-DiT-XL/1 is further improved from 1.82 to \textbf{1.65}, demonstrating that the Refiner is helpful for achieving high-fidelity generation. To see this, we present a qualitative comparison in Figure~\ref{fig:refiner_qualitative}.
The RePack decoded image preserves the overall semantic layout but exhibits reduced fine-grained texture detail.
The refined images enhance texture realism, such as the porous texture of the orange peel, while maintaining structural consistency.

\textbf{Refiner is robust w.r.t different choices of DiTs.}
A natural question is whether the improvement by Refiner is rigidly coupled with a specific choice of DiT. 
To examine this, we perform multiple learning experiments with five DiTs with different scales, from DiT-B to DiT-XL. 
As in Table~\ref{tab:refiner_backbone_scale}, the Refiner consistently improves performance on all DiT model scales.

\begin{table}[t]
\centering
\caption{\textbf{Effectiveness of the Latent-Guided Refiner across different DiT backbone scales (gFID $\downarrow$)}, evaluated after 64 training epochs.}
\label{tab:refiner_backbone_scale}
\resizebox{\linewidth}{!}{
\begin{tabular}{l c c c c c}
\toprule
\textbf{Method} & \textbf{DiT-B/1} & \textbf{DiT-B/2} & \textbf{DiT-L/2} & \textbf{DiT-XL/1} & \textbf{DiT-XL/2} \\
\midrule
RePack & 4.12 & 12.06 & 4.93 & 1.82 & 3.66 \\
RePack + Refiner & 3.99 & 11.98 & 4.83 & \textbf{1.65} & 3.20 \\
\bottomrule
\end{tabular}
}
\vspace{-5pt}
\end{table}

\begin{table}[t]
\centering
\caption{\textbf{Effectiveness of the Latent-Guided Refiner across different sampling steps (gFID $\downarrow$)}. All results are evaluated using RePack-DiT-XL/1 trained for 64 epochs.}
\label{tab:refiner_sampling_steps}
\resizebox{0.78\linewidth}{!}{ 
\begin{tabular}{l c c c}
\toprule
\textbf{Method} & \textbf{25 Steps} & \textbf{50 Steps} & \textbf{250 Steps} \\
\midrule
RePack & 2.09 & 1.87 & 1.82 \\
RePack + Refiner & 2.04 & 1.76 & \textbf{1.65} \\
\bottomrule
\end{tabular}
}
\vspace{-10pt}
\end{table}

\textbf{Application of Refiner to Other VAEs.}
It is also a question whether the Refiner can be applied to enhance a wide range of VAEs other than RePack itself. 
We find that for both VA-VAE and SD-VAE, applying Refiner leads to a decrease in rFID but an increase in gFID. We believe that the phenomenon that Refiner is most useful with RePack is related to the properties of its compact semantic manifold.
Refer to Appendix~\ref{sec:refiner_generalization} for detailed analysis.


\subsection{How the Latent Representation of RePack is Superior to Others}

\label{sec:latent_analysis}
It is known that the quality of a VAE latent representation can significantly affect the learning of DiT.  It is still unclear what the most desired properties are for an effective latent representation. Prior works suggest properties like reduced high-frequency noise \cite{se-vae} and disentangled structural information from fine-grained details \cite{dcae-1.5}.
Based on these insights, we hypothesize that an effective latent representation should preserve coherent semantic structure in low-frequency bands while retaining clean, meaningful contours in high-frequency bands. To validate this, we perform a spectral decomposition analysis on the latent codes using a 2D FFT.

\begin{figure*}[t]
  \centering
  \includegraphics[width=0.95\linewidth]{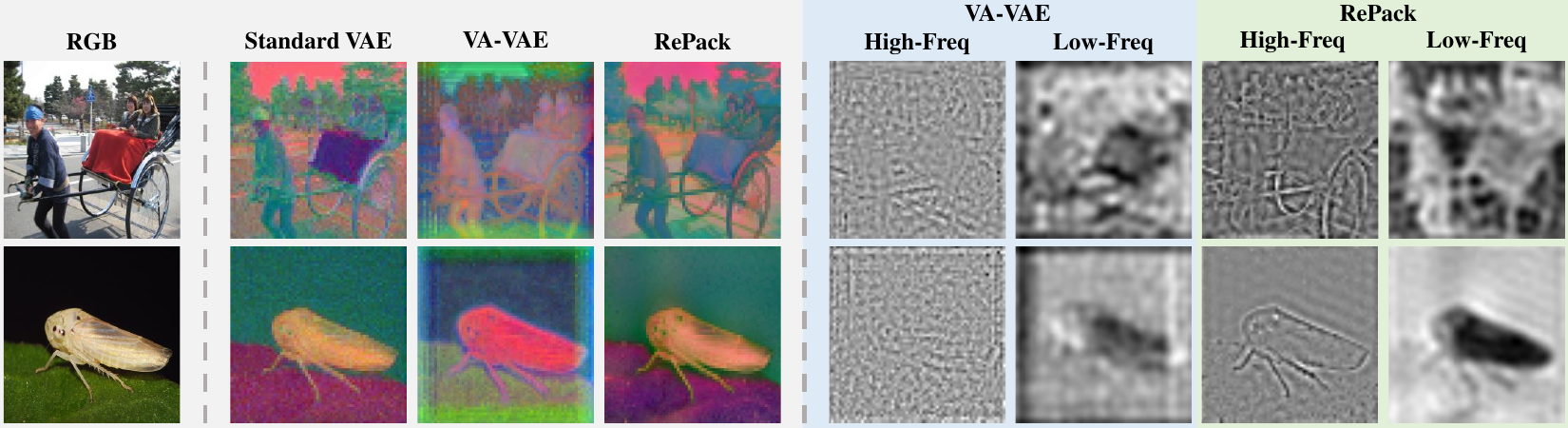}
  \caption{
  \textbf{Visual analysis of latent representations.}
  \textbf{Left:} Input images.
  \textbf{Middle:} PCA visualizations of latents from a Standard VAE (trained w/o VFM guidance), VA-VAE, and RePack.
  \textbf{Right:} Spectral decomposition of VA-VAE and RePack latents.
  }
  \label{fig:latent_vis}
\end{figure*}

As shown in Figure~\ref{fig:latent_vis}, VA-VAE exhibits strong aliasing artifacts in high-frequency components and blurred structures in low frequencies. By contrast, RePack suppresses aliasing noise while preserving distinct object contours in high frequencies and a sharp, coherent semantic layout in low frequencies.
PCA visualizations further show that RePack produces cleaner and more structured latent representations than both VA-VAE and a Standard VAE (i.e., trained from scratch without VFM supervision). 
Overall, these results indicate that RePack provides a cleaner and more disentangled latent signal. This superior representation simplifies the modeling task for DiT.
More experimental details and discussions on latent representations can be found in the Appendix \ref{sec:supp_latent_analysis}.

\subsection{Ablation Studies} 
\textbf{Impact of VFMs.}
We evaluate RePack with different VFM backbones, including DINOv2-B~\cite{DinoV2}, DINOv3-B~\cite{DinoV3}, MAE~\cite{MAE}, and SigLIP~\cite{SigLIP}, using a DiT-B/1 model. Experimental results show that RePack generalizes well across these diverse VFM architectures. In our main experiments, we adopt DINOv3-B as the default backbone to align with the latest developments in Vision Foundation Models. However, as shown in Table~\ref{tab:vfm_ablation}, DINOv2-B actually achieves the lowest gFID. 
This result shows that the effectiveness of our framework is not tightly coupled with a specific choice of VFMs. In our main results, we choose DINO-V3 because it is the latest and most widely recognized VFM. Practitioners can easily apply our framework to other VFMs. 


\begin{table}[t]
    \vspace{-10pt}
    \centering
    \caption{Generation performance (gFID) across different VFM backbones (DiT-B/1, 64 epochs).}
    \label{tab:vfm_ablation}
    \resizebox{0.65\linewidth}{!}{  
    \begin{tabular}{l c}
    \toprule
    \textbf{Backbone} & \textbf{gFID} $\downarrow$ \\
    \midrule
    MAE~\cite{MAE} & 11.66 \\
    SigLIP~\cite{SigLIP} & 5.74 \\
    DINOv3-B~\cite{DinoV3} & 4.12 \\
    \textbf{DINOv2-B}~\cite{DinoV2} & \textbf{3.84} \\
    \bottomrule
    \end{tabular}
    }
\end{table}

\begin{table}[t]
    \centering
    \caption{Ablation on Projection Layer Design (DiT-B/1, 64 epochs).}
    \label{tab:ablation_projection}
    \resizebox{0.95\linewidth}{!}{
    \begin{tabular}{l c c c}
    \toprule
    \textbf{Projection Type} & PCA & Linear (w/ Bias) & \textbf{Linear (Bias-free)} \\
    \midrule
    \textbf{gFID} $\downarrow$ & 7.56 & 5.08 & \textbf{4.12} \\
    \bottomrule
    \end{tabular}
    }
    \vspace{-7pt}
\end{table}
\textbf{Impact of Projection Layers.}
We investigate different strategies for implementing the RePack projection layer $\mathcal{P}_\theta$.
As shown in Table~\ref{tab:ablation_projection}, the learnable bias-free design is the most effective choice. 
The bias-free projection outperforms the biased variant (gFID 4.12 vs. 5.08) as it performs pure rotation and scaling without introducing unnecessary distribution shifts.
For the PCA baseline, we directly project the raw features onto the top-32 principal components calculated from the training set. However, this PCA-based variant performs the worst (gFID 7.56).
This suggests that a learnable projection is essential for capturing the semantic manifold structure for DiT training.

\textbf{Impact of Latent Dimension Sizes.}
\begin{table}[t]
    \centering
    \caption{Effect of Latent Dimension $d$ on Reconstruction and Generation Performance (DiT-B/1, 64 Epochs).}
    \label{tab:ablation_dim}
    \resizebox{0.8\linewidth}{!}{
    \begin{tabular}{c c c c c}
    \toprule
    \textbf{Dimension} ($d$) & 16 & \textbf{32 (Default)} & 64 & 768 (Raw) \\
    \midrule
    \textbf{rFID} $\downarrow$ & 1.58 & 0.83 & 0.79 & \textbf{0.74} \\
    \textbf{gFID} $\downarrow$ & \textbf{3.61} & 4.12 & 10.18 & 66.43 \\
    \bottomrule
    \end{tabular}
    }
    \vspace{-10pt}
\end{table}
We investigate the impact of the bottleneck dimension $d$ on the trade-off between reconstruction fidelity and generation efficiency.
As shown in Table~\ref{tab:ablation_dim}, reducing the dimension to $d=16$ yields the best generative performance (gFID 3.61).
This result strongly supports our core hypothesis that a more compact VFM semantic space alleviates the modeling burden for the DiT, leading to faster convergence.
However, increasing $d$ to 64 or beyond (Raw 768-dimensional features) provides diminishing returns in reconstruction (rFID) while severely degrading generation quality (gFID increases to 66.43).
While $d=16$ offers the fastest convergence, we adopt $d=32$ as our default setting to align with established baselines.
Recent work such as VA-VAE~\cite{VAVAE} and FAE~\cite{FAE} all use similar latent channels (e.g., 32).
Setting $d=32$, we strive for a fair comparison w.r.t. baselines, demonstrating that the gains of our framework are not simply a result of reducing $d$ to a lower number.

\section{Conclusion}
We present \textit{RePack then Refine}, a framework that can harness the semantic richness of VFMs without sacrificing the training efficiency of latent DiTs. The \textit{RePack} module uses a lightweight, bias-free projection to remove redundancy in raw VFM features and compress them into a compact semantic manifold. To compensate for high-frequency detail loss, we propose a \textit{Latent-Guided Refiner} that restores textures in the pixel domain while preserving latent-space structure generation.
Experiments show that \textit{RePack then Refine} achieves the fastest learning with high visual fidelity.

\textbf{Limitations and Future Work.} While our current evaluations and extensive appendix analyses sufficiently validate the core claims, exploring extended training schedules may further showcase the framework's potential. We leave scaling to large-scale tasks, such as text-to-image generation, for future work.

\clearpage

\section*{Impact Statement}
This paper presents work whose goal is to advance the field of machine learning, particularly in the area of efficient image generation. 
There are many potential societal consequences of our work, none of which we feel must be specifically highlighted here.

\bibliography{references}
\bibliographystyle{icml2026}

\newpage
\appendix
\onecolumn

\section{Model Architectures}
\label{sec:arch_details}

In this section, we provide detailed architectural specifications for the RePack framework, including the Encoder, Decoder, and the Latent-Guided Refiner.

\subsection{RePack Encoder and Decoder}
The RePack VAE consists of an asymmetric Encoder-Decoder pair. 
For the Encoder, we present the configuration that achieved the best performance in our experiments. 
Specifically, we utilize a frozen \textbf{DINOv3-ViT-B/16} backbone followed by a lightweight linear projector.
For the Decoder, we adopt the exact architecture used in \textbf{VA-VAE} \cite{VAVAE} to ensure a fair comparison and to demonstrate that our performance gains are due to the packed semantic representation rather than a stronger decoder.
The detailed configuration for both components is provided in Table~\ref{tab:vae_arch}.

\begin{table}[h]
    \centering
    \caption{\textbf{Detailed Architecture of RePack VAE.} The input resolution is $256 \times 256$. The Encoder compresses images into a $32 \times 16 \times 16$ latent space using a frozen DINOv3 backbone. The Decoder, adopted from VA-VAE~\cite{VAVAE}, reconstructs the image from this compact manifold.}
    \label{tab:vae_arch}
    \begin{small}
    \begin{tabular}{l c c l}
        \toprule
        \textbf{Module} & \textbf{Resolution} & \textbf{Channels} & \textbf{Layer Details} \\
        \midrule
        \rowcolor{gray!10} \multicolumn{4}{c}{\textit{\textbf{Encoder} (Frozen Backbone + Projector)}} \\
        \midrule
        Input Image & $256 \times 256$ & 3 & -- \\
        ViT Backbone & $16 \times 16$ & 768 & Frozen DINOv3-B/16 (Patch Features) \\
        Projector & $16 \times 16$ & $64$ & Conv2d $1\times1$, bias=False \\
        Latent Sample & $16 \times 16$ & $32$ & Gaussian Sampling (Mean/Std split) \\
        \midrule
        \rowcolor{gray!10} \multicolumn{4}{c}{\textit{\textbf{Decoder} (Adopted from VA-VAE \cite{VAVAE})}} \\
        \midrule
        Latent Input & $16 \times 16$ & 32 & Input $z$ \\
        Conv In & $16 \times 16$ & 512 & Conv2d $3\times3$ \\
        Middle Block & $16 \times 16$ & 512 & ResBlock $\to$ Attention $\to$ ResBlock \\
        Upsample Block 1 & $32 \times 32$ & 512 & ResBlock $\times 3$ $\to$ Upsample \\
        Upsample Block 2 & $64 \times 64$ & 256 & ResBlock $\times 3$ $\to$ Upsample \\
        Upsample Block 3 & $128 \times 128$ & 256 & ResBlock $\times 3$ $\to$ Upsample \\
        Upsample Block 4 & $256 \times 256$ & 128 & ResBlock $\times 3$ $\to$ Upsample \\
        Output Block & $256 \times 256$ & 128 & ResBlock $\times 3$ $\to$ Norm $\to$ Swish \\
        Reconstruction & $256 \times 256$ & 3 & Conv2d $3\times3$ \\
        \bottomrule
    \end{tabular}
    \end{small}
\end{table}
\begin{table}[h]
    \centering
    \caption{\textbf{Computational Costs and Inference Speed.} 
    Measured on a single V100 GPU with Batch Size = 1 at $256 \times 256$ resolution.}
    \label{tab:complexity}
    \begin{small}
    \begin{tabular}{l c c c c}
        \toprule
        \textbf{Module} & \textbf{Params (M)} & \textbf{GFLOPs / step} & \textbf{Throughput (img/s)} & \textbf{Peak RAM (MB)} \\
        \midrule
        RePack Encoder & 85.71 & 22.35 & 75.79 & 504.77 \\
        LightningDiT-XL/1 & 675.15 & 228.86 & 35.99 & 2868.83 \\
        RePack Decoder & 41.42 & 125.82 & 50.39 & 622.74 \\
        Refiner & 55.56 & 338.18 & 19.43 & 582.72 \\
        \bottomrule
    \end{tabular}
    \end{small}
\end{table}
\subsection{Latent-Guided Refiner Architecture}
The Latent-Guided Refiner is a conditional U-Net designed to restore high-frequency details. 
It takes the channel-wise concatenation of the base image $I_{\text{base}}$ (3 channels) and the bilinearly upsampled latent features $z_{\text{up}}$ (32 channels) as input, resulting in 35 input channels.
The network follows a U-Net structure with skip connections to propagate fine-grained information from the encoder to the decoder.
The detailed configuration for the Refiner is provided in Table~\ref{tab:refiner_arch}.

\begin{table}[h]
    \centering
    \caption{\textbf{Refiner Architecture (U-Net).} 
    The model operates at a base channel width of $C=128$. 
    Skip connections concatenate features from the encoder path to the decoder path. 
    Attention blocks are applied at the $32 \times 32$ resolution to capture global context.}
    \label{tab:refiner_arch}
    \begin{small}
    \begin{tabular}{l c c l}
        \toprule
        \textbf{Stage} & \textbf{Resolution} & \textbf{Ch (In/Out)} & \textbf{Layer Composition} \\
        \midrule
        \textbf{Input} & $256 \times 256$ & $35 \to 128$ & Concat($I_{base}, z_{up}$) $\to$ Conv $3\times3$ \\
        \midrule
        \multicolumn{4}{l}{\textit{Downsampling Path (Encoder)}} \\
        Down 1 & $256 \to 128$ & $128 \to 128$ & ResBlock $\times 2$ $\to$ Downsample (Stride 2) \\
        Down 2 & $128 \to 64$ & $128 \to 256$ & ResBlock $\times 2$ $\to$ Downsample (Stride 2) \\
        Down 3 & $64 \to 32$ & $256 \to 512$ & ResBlock $\times 2$ $\to$ Attention $\to$ Downsample \\
        \midrule
        \multicolumn{4}{l}{\textit{Bottleneck}} \\
        Middle & $32 \times 32$ & $512 \to 512$ & ResBlock ($512 \!\to\! 1024$) $\to$ Attn $\to$ ResBlock ($1024 \!\to\! 512$) \\
        \midrule
        \multicolumn{4}{l}{\textit{Upsampling Path (Decoder) - Inputs include Skip Connections}} \\
        Up 1 & $32 \to 64$ & $1024 \to 512$ & Upsample $\to$ Concat $\to$ ResBlock $\to$ Attn \\
        Up 2 & $64 \to 128$ & $768 \to 256$ & Upsample $\to$ Concat $\to$ ResBlock $\times 2$ \\
        Up 3 & $128 \to 256$ & $384 \to 128$ & Upsample $\to$ Concat $\to$ ResBlock $\times 2$ \\
        \midrule
        \textbf{Output} & $256 \times 256$ & $256 \to 3$ & Concat(Skip) $\to$ ResBlock $\to$ Norm $\to$ SiLU $\to$ Conv $\to$ Tanh \\
        \bottomrule
    \end{tabular}
    \end{small}
\end{table}

\subsection{Computational Complexity and Inference Efficiency}
\label{sec:complexity}

To evaluate the practical deployment costs of RePack, we analyze the computational complexity and memory footprint of each component.
Table~\ref{tab:complexity} reports the number of parameters, GFLOPs, inference throughput (img/s), and peak memory usage.
All metrics are measured on a single \textit{NVIDIA Tesla V100} GPU with a batch size of 1 to simulate a real inference scenario.

\paragraph{Analysis of Refiner Complexity.}
As shown in Table~\ref{tab:complexity}, the RePack Encoder and Decoder are efficient, maintaining high throughput with low memory consumption.
We observe that the \textit{Refiner} exhibits a higher GFLOPs count (338.18 G) compared to the VAE components.
This is expected as the Refiner operates directly on high-resolution pixel inputs ($256 \times 256$) with 35 channels, whereas the VAE components process data in a compressed latent space.
However, it is crucial to note that unlike the Diffusion Transformer (DiT), which requires tens or hundreds of iterative function evaluations during sampling, the Refiner is a single-step feed-forward network.
Consequently, despite the higher GFLOPs, it achieves a throughput of \textit{19.43 img/s}, which is sufficient for most generation applications.
In future work, we plan to further enhance the Refiner's efficiency by applying model compression techniques, such as pruning and quantization, to reduce computational overhead and further boost throughput.

\section{Related Works}
\label{sec:related_works}

\subsection{VFM-Driven Latent Representations}
\label{sec:rw_vfm_driven}

\textbf{Direct High-Dimensional Utilization.}
Some methods bypass encoder training by using raw VFM features directly as latents.
\textbf{RAE}~\cite{RAE} and \textbf{T2I-RAE}~\cite{T2I-RAE} utilize frozen features from models like DINOv2, shifting the reconstruction burden to the decoder to maximize semantic richness.
Similarly, \textbf{SVG}~\cite{SVG} augments frozen DINO features with a lightweight residual branch. This strategy is subsequently scaled to text-to-image synthesis in \textbf{SVG-T2I}~\cite{SVG-T2I}.

\textbf{Adapted and Compressed VFM Latents.}
Other approaches project VFM features into lower-dimensional spaces to reduce redundancy.
\textbf{FAE}~\cite{FAE} employs an attention encoder to compress high-dimensional features into compact latents.
\textbf{VFM-VAE}~\cite{VFM-VAE} integrates frozen VFMs with a multi-scale fusion decoder to bridge semantic and pixel domains.
\textbf{Send-VAE}~\cite{Send-VAE} aligns VAE latents with VFM hierarchies to enhance disentanglement.
\textbf{PS-VAE}~\cite{PS-VAE} maps frozen features into a KL-regularized space and jointly optimizes for pixel and semantic consistency to preserve fine-grained textures.

While these methods effectively leverage VFM semantics, direct utilization approaches (e.g., RAE) exhibit slower convergence, and adaptation methods (e.g., FAE, PS-VAE) often involve more complex nonlinear encoders.
In contrast, \textbf{RePack} explores a lightweight alternative by identifying a low-dimensional structure within VFM features and applying a bias-free linear projection. This design reduces redundancy while avoiding additional architectural complexity.

\subsection{Representation Alignment and Distillation}
\label{sec:rw_alignment}

Instead of modifying the latent space architecture, some methods enforce alignment during training via auxiliary losses.
\textbf{REPA}~\cite{REPA} regularizes diffusion models by aligning noisy states with clean VFM representations.
This paradigm is subsequently extended to end-to-end VAE-DiT tuning in \textbf{REPA-E}~\cite{REPA-E}, U-Net architectures in \textbf{U-REPA}~\cite{U-REPA}, and video generation in \textbf{VideoREPA}~\cite{VideoREPA}.
\textbf{REG}~\cite{REG} takes a different approach by explicitly entangling high-level class tokens with image latents to guide the denoising process.
\textbf{VA-VAE}~\cite{VAVAE} introduces a VFM alignment loss to regularize the VAE latent space towards semantic distributions without changing the inference architecture.

Alignment strategies improve convergence by introducing auxiliary regularization, while \textbf{RePack} follows a different design choice and explicitly constructs the latent space from VFM features via projection. This design avoids additional loss terms and allows the DiT to operate directly on a semantically structured manifold.

\subsection{Structural and Spectral Regularization}
\label{sec:rw_structural}
Some methods accelerate training by optimizing the tokenizer’s internal properties through mathematical or structural constraints, without relying on external VFM guidance.
\textbf{EQ-VAE}~\cite{EQ-VAE} encourages equivariance to spatial transformations such as rotation and scaling to smooth the latent manifold.
\textbf{SE}~\cite{se-vae} regularizes spectral properties to enable a coarse-to-fine spectral autoregression.
\textbf{FA-VAE}~\cite{FA-VAE} utilizes wavelet decomposition to explicitly decouple high-frequency details from low-frequency structure.
\textbf{HieraTok}~\cite{HieraTok} employs a multi-scale ViT with scale-causal attention to capture hierarchical detail progressively.
\textbf{DeToK}~\cite{DeToK} aligns the tokenizer’s latent space with the downstream denoising objective by training the decoder to reconstruct clean images from corrupted latents.
These approaches rely on auxiliary constraints to shape the VAE latent space, whereas \textbf{RePack} directly leverages the pre-aligned semantic structure of VFMs through projection onto a compact manifold.

\subsection{Internal Guidance and Self-Evolution}
\label{sec:rw_internal}

Recent works also explore accelerating convergence by leveraging the model's own internal dynamics rather than external teachers.
\textbf{SRA}~\cite{SRA} aligns weaker early-layer representations with stronger deep-layer ones to provide internal guidance.
\textbf{IG}~\cite{IG} utilizes intermediate layer supervision to stabilize gradients and employs self-guidance during sampling to improve quality.
\textbf{Self-Transcendence}~\cite{Self-Transcendence} aligns shallow features with VAE latents and then applies internal classifier-free guidance to amplify semantic richness.
These methods optimize the training trajectory through internal distillation, while \textbf{RePack} operates at the level of input representation.

\subsection{Unified Frameworks and Optimization Dynamics}
\label{sec:rw_unified}

Recently, several studies focus on unifying perception and generation or optimizing training dynamics to accelerate convergence.
\textbf{MingTok}~\cite{MingTok} unifies image understanding and generation within a continuous autoregressive framework to eliminate quantization errors.
\textbf{GloTok}~\cite{GloTok} captures global semantic relationships via codebook-wise histogram learning.
\textbf{Xu \textit{et al.}}~\cite{Optimal-Loss} propose an analytical loss gap metric to diagnose underfitting at specific noise levels, enabling a principled training schedule that prioritizes harder steps.
While these works focus on architectural design or loss formulation, \textbf{RePack} instead focuses on the efficiency of the latent representation itself.

\subsection{Related Methods for the Latent-Guided Refiner}
\label{sec:rw_refiner}

Our Latent-Guided Refiner draws inspiration from Pixel Diffusion Models (PDMs) such as \textbf{DeCo}~\cite{DeCo} and \textbf{PixelDiT}~\cite{PixelDiT}, which demonstrate the effectiveness of decoupling high-frequency texture modeling from low-frequency semantic generation.
Beyond direct generation, refinement techniques are widely adopted to enhance quality or restore details.
\textbf{SDEdit}~\cite{SDEdit} utilizes a stochastic process that adds noise to a coarse guide and iteratively denoises it to recover realism. \textbf{BetterDepth}~\cite{BetterDepth} applies a similar diffusion-based refinement strategy to enhance fine-grained details in zero-shot depth estimation.
To improve alignment and fidelity, \textbf{OmniRefiner}~\cite{OmniRefiner} employs Group Relative Policy Optimization to align outputs with high-resolution reference patches, a direction further refined by \textbf{Chunk-GRPO}~\cite{Chunk-GRPO}, which optimizes generation at the temporal chunk level to fix advantage attribution.
\textbf{Q-Refine}~\cite{Q-Refine} introduces a quality-aware pipeline using pixel-level Image Quality Assessment (IQA) maps to strictly guide the restoration of local defects.
In specific restoration domains such as blind face restoration, methods often leverage strong external priors. 
For example, \textbf{GFP-GAN}~\cite{GFP-GAN} utilizes pre-trained GAN priors to recover textures, while \textbf{CodeFormer}~\cite{CodeFormer} maps degraded inputs to a discrete codebook to reduce restoration uncertainty.
In contrast to these approaches, RePack uses the packed latent $z$ as a structural reference. The Refiner operates as a deterministic, offline module and focuses on complementing texture details, while keeping the overall inference pipeline unchanged.

\section{Detailed Configurations}
\label{ref:detail_config}
We provide the training hyperparameters for three stages of the RePack framework: the Representation Packing (VAE), the Generative Model (DiT), and the Latent-Guided Refiner.

\subsection{Stage 1: Representation Packing (VAE)}
\begin{table}
  \centering
  \caption{Essential training hyperparameters for RePack Stage 1.}
  \label{tab:stage1_hyperparams}
  
  \setlength{\aboverulesep}{0pt} 
  \setlength{\belowrulesep}{0pt}
  \renewcommand{\arraystretch}{1.12}

  \resizebox{0.5\linewidth}{!}{ 
    \begin{tabular}{lc|lc}
    \toprule
    \textbf{Configuration} & \textbf{Value} & \textbf{Loss Weights} & \textbf{Value} \\
    \midrule
    Base Learning Rate & $1.0 \times 10^{-5}$ & LPIPS weight ($\lambda_{\text{lpips}}$) & 0.2 \\
    Batch Size & 32 & Adversarial weight ($\lambda_{\text{adv}}$) & 0.5 \\
    Embed Dimension & 32 & FFL weight ($\lambda_{\text{f}}$) & 0.05 \\
    Resolution & $256 \times 256$ & Watson weight ($\lambda_{\text{W}}$) & 0.005 \\
    \bottomrule
    \end{tabular}
  }
\end{table}
The Stage 1 training aims to compress high-dimensional features from a frozen VFM (DINOv3-B/16) into a compact 32-dimensional semantic manifold. By incorporating Focal Frequency Loss \cite{focal-freq-loss} and Watson Perceptual's Loss supervision \cite{watson-loss}, the reconstruction quality is enhanced, improving the rFID from 0.8789 to 0.8295. The essential training parameters are summarized in Table~\ref{tab:stage1_hyperparams}.

\subsubsection{Focal Frequency Loss (FFL)}
We employ Focal Frequency Loss (FFL) \cite{focal-freq-loss} to adaptively focus on hard-to-reconstruct frequency components. Given the frequency spectra of the real image $F_r(u,v)$ and the reconstructed image $F_f(u,v)$ obtained via 2D DFT, the loss is defined as:
$$ \mathcal{L}_{\text{Focal}} = \frac{1}{MN} \sum_{u=0}^{M-1} \sum_{v=0}^{N-1} w(u, v) \left| F_r(u, v) - F_f(u, v) \right|^2 $$
The spectrum weight matrix $w(u,v)$ is dynamically updated to down-weight easy frequencies and emphasize hard ones:
$$ w(u, v) = \left| F_r(u, v) - F_f(u, v) \right|^\alpha $$
In our implementation, we set $\alpha = 1.0$ as a balanced trade-off between focus and stability.

\subsubsection{Watson Perceptual Loss}
Watson Perceptual Loss \cite{watson-loss} aligns the reconstruction with human visual perception. It accounts for luminance masking and contrast masking by normalizing frequency errors with a visibility threshold $S_{ijk}$:
$$ \mathcal{L}_{\text{W}} = \sqrt[p]{\sum_{i, j, k} \left| \frac{C_{ijk} - C'_{ijk}}{S_{ijk}} \right|^p} ,$$
where $C$ and $C'$ are transform coefficients. This approach prevents the model from being over-penalized by visually imperceptible differences in bright or high-contrast regions, thus preserving more realistic textures in the final reconstruction.

\subsection{Stage 2: Generative Modeling (DiT)}
For the generative stage, we adopt LightningDiT \cite{VAVAE}, an optimized DiT variant designed for fast convergence and high-fidelity generation. LightningDiT incorporates architectural improvements such as SwiGLU FFN, RMSNorm, and Rotary Positional Embeddings (RoPE).

\subsubsection{Model and Loss Configurations}
The model architecture and training loss parameters are detailed in Table~\ref{tab:dit_arch_loss}.

\begin{table}[H]
    \centering
    \caption{LightningDiT-XL/1 Architecture and Loss configurations.}
    \label{tab:dit_arch_loss}
    \small 
    \setlength{\aboverulesep}{0pt} 
    \setlength{\belowrulesep}{0pt} 
    \renewcommand{\arraystretch}{1.1} 
    
    \begin{tabular}{ll|ll}
    \toprule
    \textbf{Parameter} & \textbf{Value} & \textbf{Parameter} & \textbf{Value} \\
    \midrule
    \rowcolor{gray!10} \multicolumn{2}{l|}{\textbf{Architecture}} & \multicolumn{2}{l}{\textbf{Training / Loss}} \\
    Input Channels     & 32      & Path Type         & Linear (Rectified Flow) \\
    FFN Type           & SwiGLU  & Prediction Target & Velocity                \\
    Normalization      & RMSNorm & Use Logit-Normal  & True                    \\
    Position Embedding & RoPE    & Use Cosine Loss   & True                    \\
                       &         & AdamW $\beta_2$   & 0.95                    \\
    \bottomrule
    \end{tabular}
\end{table}

\begin{wraptable}{r}{0.45\textwidth} 
\centering
\caption{FID-50K performance comparison for DiT-XL/1: LR Scheduler vs. Manual Stage-wise Drop.}
\label{tab:lr_strategy_comparison}
\resizebox{\linewidth}{!}{ 
\begin{tabular}{lccccc}
\toprule
\textbf{Epoch} & 16 & 32 & 48 & 56 & \textbf{64} \\
\midrule
Standard Scheduler & 3.31 & 2.53 & 2.10 & 1.99 & 1.98 \\
\textbf{Manual Stage-wise} & 3.31 & 2.48 & 2.24 & 1.87 & \textbf{1.82} \\
\bottomrule
\end{tabular}
}
\end{wraptable}

\subsubsection{Stage-wise Training Strategy}
We found that a manual stage-wise training strategy, rather than a continuous learning rate scheduler, leads to better convergence and final generative quality. As shown in the Figure \ref{fig:fitting_speed}, we observed a sharp improvement in FID after a manual learning rate adjustment at Epoch 48.

Specifically, we initially train the model with a learning rate of $2.0 \times 10^{-4}$. At Epoch 48, the learning rate is dropped to $1.0 \times 10^{-4}$. This approach allows the model to first capture global semantic structures and subsequently refine details within the compact manifold. As shown in Table~\ref{tab:lr_strategy_comparison}, while a standard scheduler achieves a respectable FID of 1.98 at Epoch 64, our manual stage-wise strategy reaches a superior FID of \textbf{1.82}.
\paragraph{Scalability of the Stage-wise Strategy.}
It is worth noting that while the manual learning rate decay yields significant gains for the large-scale DiT-XL/1, this benefit does not generalize to smaller backbones. Experimental results show that applying the same sharp decay strategy to DiT-B/1 results in a gFID of 5.08, which is inferior to the 4.12 achieved with the standard scheduler. Similarly, DiT-B/2 exhibits a slight performance degradation (12.19 vs. 12.06).

\begin{wraptable}{r}{0.48\textwidth} 
\centering
\vspace{-15pt} 
\caption{Sampling and Inference Hyperparameters.}
\label{tab:sampling_params}
\small 
\begin{tabular}{ll}
\toprule
\textbf{Parameter} & \textbf{Value} \\
\midrule
Sampling Method    & Euler (ODE Solver) \\
Sampling Steps     & 250                \\
ODE Tolerance (atol/rtol) & $10^{-6}$ / $10^{-3}$ \\
\midrule
CFG Scale ($\omega$) & 15.0              \\
CFG Interval Start & 0.11               \\
Timestep Shift     & 0.0                \\
FID Sample Count   & 50,000             \\
\bottomrule
\end{tabular}
\end{wraptable}

\subsubsection{Numerical Integration and Sampling}
We attribute this behavior to differences in model capacity and optimization dynamics. Larger models can capture global semantics early and benefit from a sharp learning rate drop for refinement, 
whereas smaller models require longer exploration with a higher learning rate. As a result, stage-wise decay is effective for large models, while standard scheduling remains more reliable for lightweight backbones.

We evaluate the generative performance of RePack by solving the deterministic Probability Flow Ordinary Differential Equation (ODE) associated with the learned velocity field. Following the Rectified Flow formulation, we utilize a first-order Euler integrator for numerical solution. 

To achieve high-fidelity generation and precise class alignment, we employ Classifier-Free Guidance (CFG) with a scale of 15.0. To further optimize the balance between perceptual quality and sample diversity, we incorporate a CFG interval strategy starting at 0.11. Consistent with our objective of a fair comparison with established baselines, we maintain a timestep shift of 0.0 during the sampling process. The full sampling configuration is summarized in Table~\ref{tab:sampling_params}.

\subsubsection{Stage 3: Latent-Guided Refiner}
The Latent-Guided Refiner bridges the gap between latent-domain semantics and pixel-domain high-frequency fidelity. It serves as a module that restores fine-grained textures by using compact 32-dimensional latents as structural guidance.
We employ a PatchGAN \cite{Patch-GAN} discriminator for the adversarial loss $\mathcal{L}_{\text{adv}}$ to encourage the synthesis of realistic high-frequency textures such as sharp edges and fine fur patterns. Core hyperparameters are summarized in Table~\ref{tab:refiner_hyperparams}.
\begin{wraptable}{r}{0.48\textwidth} 
\centering
\vspace{25pt}
\caption{Essential Hyperparameters for the Latent-Guided Refiner.}
\label{tab:refiner_hyperparams}
\small 
\begin{tabular}{lc}
\toprule
\textbf{Configuration} & \textbf{Value} \\
\midrule
Learning Rate          & $1.0 \times 10^{-4}$ \\
Batch Size             & 64 \\
Latent Dimension ($z$) & 32 \\
Input Channels          & 35 \\
LPIPS weight ($\lambda_{\text{lpips}}$) & 1.0 \\
GAN weight ($\lambda_{\text{adv}}$)     & 0.5 \\
\bottomrule
\end{tabular}
\end{wraptable}

\section{Additional Experimental Results}

In this section, we provide further results and analysis to validate the robustness and generative performance of RePack.

\subsection{Performance without Classifier-Free Guidance}
We evaluate RePack on ImageNet $256 \times 256$ without classifier-free guidance. As shown in Table~\ref{tab:no_cfg_results}, RePack achieves a gFID of 3.81 after only 64 training epochs, outperforming VA-VAE (5.14) and the original DiT (9.62). This demonstrates that the compact 32-dimensional semantic manifold provides a cleaner and more effective optimization target, enabling rapid convergence even without external guidance. Although methods such as FAE and RAE achieve lower FID with longer training or larger VAEs, RePack’s strong early-stage performance highlights its advantage for efficient training and fast iteration.

\begin{table}[h]
\centering
\caption{Class-conditional Image Generation Results \textbf{without CFG} for DiT-XL/1 on ImageNet $256 \times 256$.}
\label{tab:no_cfg_results}
\resizebox{0.5\linewidth}{!}{
\begin{tabular}{lccc}
\toprule
\textbf{Method} & \textbf{Epochs} & \textbf{gFID $\downarrow$} & \textbf{IS $\uparrow$} \\
\midrule
\rowcolor{gray!10}\textbf{AR Models} & & & \\
LlamaGen \cite{LlamaGen} & 300 & 9.38 & 112.9 \\
MAR \cite{MAR} & 800 & 2.35 & 227.8 \\
\midrule
\rowcolor{gray!10}\textbf{LDMs} & & & \\
DiT \cite{DiT} & 1400 & 9.62 & 121.5 \\
SiT \cite{SiT} & 1400 & 8.61 & 131.7 \\
VA-VAE \cite{VAVAE} & 64 & 5.14 & 130.2 \\
VA-VAE \cite{VAVAE} & 800 & 2.17 & 205.6 \\
FAE \cite{FAE} & 64 & 2.55 & 189.9 \\
RAE (DiT-XL) \cite{RAE} & 800 & 1.87 & 209.7 \\
\midrule
\rowcolor{highlightblue} \textbf{RePack} & \textbf{64} & \textbf{3.81} & \textbf{151.4} \\
\rowcolor{highlightblue} \textbf{RePack + Refiner} & \textbf{64} & \textbf{3.79} & \textbf{152.5} \\
\bottomrule
\end{tabular}
}
\end{table}

\subsection{Extended Analysis of Latent Representations}
\label{sec:supp_latent_analysis}

In this section, we provide the mathematical formulation for the spectral decomposition analysis presented in the main text (Section \ref{sec:latent_analysis}) and offer a more detailed discussion on the structural properties of the learned latent spaces.

\begin{table*}[t]
\centering
\caption{\textbf{Taxonomy of Latent Space Construction Paradigms.} We categorize recent approaches based on their utilization of Vision Foundation Models (VFMs) and training strategies.}
\label{tab:latent_space_paradigms}
\resizebox{\textwidth}{!}{
\begin{tabular}{l p{4.5cm} p{4.2cm} p{4.2cm} p{4.2cm}}
\toprule
\textbf{Category} & \textbf{Representative Methods} & \textbf{Core Mechanism} & \textbf{Key Advantages} & \textbf{Trade-offs} \\
\midrule
\textbf{I. Native Structural} & EQ-VAE~\cite{EQ-VAE}, \newline SE~\cite{se-vae}, \newline DCAE-1.5~\cite{dcae-1.5} & 
Optimizes latent structure via internal constraints (e.g., equivariance, spectral regularity) without VFM guidance. & 
Self-contained training; effective disentanglement of structure and detail. & 
Lacks high-level semantic priors, resulting in slower DiT convergence compared to VFM-assisted methods. \\
\midrule
\textbf{II. VFM-Aligned} & VA-VAE~\cite{VAVAE}, \newline REPA-E~\cite{REPA-E} & 
Indirectly leverages VFMs to regularize VAE training via alignment losses. & 
Produces a structured latent space that balances semantics and reconstruction. & 
Increases the complexity and computational cost of the VAE training stage. \\
\midrule
\textbf{III. Direct VFM} & RAE~\cite{RAE}, \newline SVG~\cite{SVG} & 
Directly utilizes frozen features from pre-trained VFMs (e.g., DINO, CLIP) as latents. & 
Maximal semantic richness; bypasses expensive encoder training by utilizing frozen VFM features. & 
High dimensionality leads to feature redundancy; overlooks the encoder's fundamental role in information compression. \\
\midrule
\textbf{IV. Pixel / PDM} & DeCo~\cite{DeCo}, \newline JiT~\cite{JiT}, \newline PixelDiT~\cite{PixelDiT} & 
Models semantics via patchified inputs and restores high-freq details using specialized components. & 
End-to-end training; decouples semantics from texture. & 
Slower convergence speed; requires dedicated high-resolution learning modules. \\
\bottomrule
\end{tabular}
}
\vspace{-10pt}
\end{table*}

\subsubsection{Spectral Decomposition}
To analyze the frequency characteristics, we decompose the latent map $z$ into low- and high-frequency components via spectral filtering. We define a binary low-pass mask $M_r$ based on a normalized cutoff $r \in [0, 1]$:
\begin{equation}
M_r(u, v) = \mathbf{1}\left[ \sqrt{(u-u_0)^2 + (v-v_0)^2} \le r \cdot \min(h, w)/2 \right],
\end{equation}
where $(u_0, v_0)$ is the spectral center. The decomposition is then computed via the 2D FFT ($\mathcal{F}$) and its inverse ($\mathcal{F}^{-1}$):
\begin{equation}
z_{\text{low}} = \mathcal{F}^{-1} \left( \mathcal{F}(z) \odot M_r \right), \quad z_{\text{high}} = \mathcal{F}^{-1} \left( \mathcal{F}(z) \odot (1 - M_r) \right).
\end{equation}
In our experiments, we set $r=0.25$ to separate the primary semantic layout from high-frequency details.

\subsubsection{Principal Component Analysis (PCA)}
To visualize the latent structure, we apply PCA to a single latent feature map $z \in \mathbb{R}^{d \times h \times w}$.
We flatten $z$ into $N$ feature vectors ($N=h \cdot w$) and project them onto the top-3 principal components to obtain a compressed representation $Y \in \mathbb{R}^{N \times 3}$.
These three components are then channel-wise min-max normalized to $[0, 1]$, reshaped to $(h, w, 3)$, and assigned to the R, G, and B channels to form the final visualization.

\subsubsection{Positioning RePack: Synergy over Competition}
As illustrated in Table~\ref{tab:latent_space_paradigms}, each paradigm offers distinct advantages and trade-offs.
For instance, Category II methods like VA-VAE have been successfully validated in large-scale production models (e.g., HunyuanImage-2.1 \cite{HunyuanImage-2.1}), proving the effectiveness of aligned representations.
It is worth noting that the analysis in Section \ref{sec:latent_analysis} is not intended to dismiss other methods, but rather to understand the implications of different design choices from the perspective of representation structure.
Building upon these insights, our work explores a complementary direction: \textbf{how can we synthesize the strengths of these distinct paradigms?}

The spectral analysis in Section \ref{sec:latent_analysis} (Main Text) reveals our design philosophy.
RePack integrates the foundational principles of these paradigms into a unified framework:
\begin{enumerate}
    \item \textbf{Insight from Cat. II (Compression):} We follow the common view that compressing the latent manifold helps simplify the modeling task. RePack adopts this idea by projecting features into a compact low-dimensional space ($d=32$).
    \item \textbf{Insight from Cat. III (Semantic Fidelity):} We observe that directly utilizing signals from frozen VFMs preserves \textit{higher semantic purity} compared to distilling them into a complex learnable encoder.
    \item \textbf{Insight from Cat. IV (Decoupling):} We follow a frequency-decoupled design. High-frequency details are handled by a dedicated Refiner.
\end{enumerate}
RePack's experimental results demonstrate that synthesizing the compression mechanism of VAEs, the semantic purity of frozen VFMs, and the decoupling strategy of PDMs provides a highly efficient solution for generative modeling.

\subsection{Extended Analysis of Representation Redundancy}
\label{app:redundancy_analysis}

We provide more detailed mathematical formulation and further analysis of the redundancy analysis presented in Section \ref{sec:redundancy_analysis}.

\subsubsection{Spectral Analysis Formulation}
Let $\mathcal{D} = \{x_1, \dots, x_N\}$ be a subset of images sampled from ImageNet-1K. We extract the raw feature tokens using the frozen VFM encoder $\mathcal{E}_\phi$. Let $\mathbf{z}_i \in \mathbb{R}^D$ denote the flattened feature vector for image $x_i$, where $D=768$ for ViT-B.
We construct the centered feature matrix $\mathbf{Z} \in \mathbb{R}^{M \times D}$, where $M$ is the total number of tokens across the dataset. The covariance matrix is given by:
\begin{equation}
    \mathbf{C} = \frac{1}{M-1} \mathbf{Z}^\top \mathbf{Z}.
\end{equation}
We perform the eigen-decomposition of $\mathbf{C}$ such that $\mathbf{C}\mathbf{v}_j = \lambda_j \mathbf{v}_j$, where $\lambda_1 \geq \lambda_2 \geq \dots \geq \lambda_D \geq 0$ are the eigenvalues representing the variance along each principal component.

\subsubsection{Effective Dimensionality}
To quantify the information content, we define the Cumulative Explained Variance (CEV) ratio at dimension $k$ as:
\begin{equation}
    R(k) = \frac{\sum_{j=1}^k \lambda_j}{\sum_{j=1}^D \lambda_j}.
\end{equation}
Our empirical evaluation on DINOv3 features yields the following observations:
\begin{itemize}
    \item \textbf{Significant Sparsity:} At $k=32$, we observe $R(32) \approx 0.77$. This implies that a subspace utilizing only $4.1\%$ of the feature dimensions captures nearly $80\%$ of the semantic energy.
    \item \textbf{Heavy-Tail Distribution:} The eigenvalue spectrum exhibits a sharp decay followed by a long tail, where $\lambda_j$ for $j > 32$ corresponds to marginal variance contributions (approx. $23\%$), largely attributed to high-frequency texture noise.
\end{itemize}

In contrast, the packed representation (RePack) exhibits a near-linear CEV curve relative to its normalized dimensions. This indicates a more uniform distribution of variance across the projected channels, demonstrating that RePack maximizes information entropy within the compact manifold and effectively eliminates inter-channel redundancy.

\subsubsection{Alignment with Linear Projection}
\label{app:PCA_analysis}
The validity of using PCA to motivate our design lies in the architecture of RePack itself. The RePack Projector $\mathcal{P}_\theta$ is defined as a learnable linear transformation $W \in \mathbb{R}^{D \times d}$.
According to the Eckart-Young-Mirsky theorem \cite{Eckart-Young-Mirsky}, the optimal rank-$d$ linear approximation of the matrix $\mathbf{Z}$ (in the Frobenius norm sense) is the truncation of its Singular Value Decomposition (SVD) to the top-$d$ singular values.

Therefore, the PCA result provides the theoretical upper bound for the information capacity of our linear projector. The observed \textit{elbow} around $d=32$ serves as the optimal operating point that balances semantic retention ($R(32) \approx 0.77$) against the computational efficiency of the subsequent DiT training.

\subsection{Linear Probing Evaluation}
\label{app:linear_probing}
To assess the semantic discriminability of the learned latent representations, we conduct a linear probing evaluation on the ImageNet-1K benchmark. We freeze the pre-trained encoder and train a supervised linear classifier on top of the extracted features.

\subsubsection{Experimental Setup}
Given an input image $x$, let $Z = \mathcal{E}(x) \in \mathbb{R}^{d \times h \times w}$ denote the frozen feature map. We define the linear probing objective as mapping the spatially averaged features directly to class logits $\mathbf{l} \in \mathbb{R}^C$:
\begin{equation}
\mathbf{l} = \mathbf{W} \left( \frac{1}{h w} \sum_{i=1}^{h} \sum_{j=1}^{w} Z_{:, i, j} \right) + \mathbf{b},
\end{equation}
where the term in parentheses represents the Global Average Pooling (GAP) operation. $\mathbf{W} \in \mathbb{R}^{C \times d}$ and $\mathbf{b} \in \mathbb{R}^C$ are the learnable weights and bias of the linear classifier ($C=1000$), optimized via standard cross-entropy loss while keeping $\mathcal{E}$ frozen.
The parameters $\{\mathbf{W}, \mathbf{b}\}$ are optimized while keeping $\mathcal{E}$ frozen.
The linear classifier is trained for 200 epochs using SGD with a momentum of 0.9 and no weight decay. 
We employ a Cosine Annealing learning rate schedule, initialized at 0.1 and decaying to 0.0001. 
All experiments are performed using a batch size of 2048.

\subsubsection{Results and Analysis}
Table \ref{tab:linear_probing_full} reports the detailed performance metrics, including Training Accuracy, Validation Accuracy, and the number of trainable parameters in the linear probe.

\begin{table*}[h]
    \centering
    \caption{\textbf{Linear Probing Performance on ImageNet-1K.} Note that the classifier for DINOv3-B has \textbf{24$\times$} more parameters compared to the RePack variants.}
    \label{tab:linear_probing_full}
    \resizebox{0.8\textwidth}{!}{
    \begin{tabular}{l c c c c c}
    \toprule
    \multirow{2}{*}{\textbf{Method}} & \textbf{Input Feature} & \textbf{Linear Layer} & \textbf{Compression} & \textbf{Train Acc.} & \textbf{Val Acc.} \\
    & \textbf{Shape} ($d$) & \textbf{Params} (M) & \textbf{Ratio} & \textbf{(\%)} & \textbf{(\%)} \\
    \midrule
    \textbf{DINOv3-B-RePack} & $1 \times 32$ & $\mathbf{0.03}$ & $\mathbf{24\times}$ & 45.24 & \textbf{42.48} \\
    PCA-RePack & $1 \times 32$ & $0.03$ & $24\times$ & 58.68 & 56.13 \\
    VA-VAE~\cite{VAVAE} & $1 \times 32$ & $0.03$ & $24\times$ & 31.00 & 28.43 \\
    Standard VAE & $1 \times 32$ & $0.03$ & $24\times$ & 0.83 & 0.85 \\
    \midrule
    \color{gray} DINOv3-B & \color{gray} $1 \times 768$ & \color{gray} $0.77$ & \color{gray} $1\times$ & \color{gray} 83.21 & \color{gray} 82.56 \\
    \bottomrule
    \end{tabular}
    }
\end{table*}

The results highlight three key observations:
\begin{enumerate}
    \item \textbf{Superiority over Baselines:} RePack achieves a validation accuracy of \textbf{42.48\%}, outperforming VA-VAE (28.43\%) and the Standard VAE (0.85\%). This confirms that our projection-based compression effectively retains core semantic structures that are often lost in standard reconstruction-based VAE training.
    
    \item \textbf{Efficiency Trade-off:} While raw DINOv3 features achieve 82.56\% accuracy, this comes at the cost of a 768-dimensional space. The linear classifier for DINOv3 requires \textbf{0.77M} parameters, whereas RePack's classifier uses only \textbf{0.03M}. Despite a $\mathbf{24\times}$ reduction in dimensionality and classifier capacity, RePack maintains resonable separability.
    
    \item \textbf{Reference to PCA:} PCA-RePack (56.13\%) can be regarded as the theoretical upper bound of linear separability within this 32-dimensional subspace, as justified in Appendix~\ref{app:PCA_analysis} based on the Eckart–Young–Mirsky theorem.
    The performance gap between RePack (42.48\%) and PCA arises from an intentional design trade-off. 
    Unlike PCA, which is optimized to maximize variance, RePack compromises some degree of semantic separability in order to satisfy the spatial reconstruction constraints required for image generation.
    Nevertheless, extensive generative experiments demonstrate that the semantic information preserved by RePack remains stable and is fully sufficient to support high-quality DiT synthesis.
\end{enumerate}

\subsubsection{Generalization of the Refiner to Other Latent Spaces}
\label{sec:refiner_generalization}
\begin{wraptable}{r}{0.6\textwidth} 
\centering
\vspace{-18pt} 
\caption{\textbf{Effectiveness of the Latent-Guided Refiner across different latent spaces.} While the Refiner consistently improves reconstruction (rFID), it degrades generation performance (gFID) for VA-VAE and SD-VAE, unlike RePack.}

To investigate the generalization of the Latent-Guided Refiner, we apply it to VA-VAE~\cite{VAVAE} and the standard SD-VAE~\cite{DiT}.

\label{tab:refiner_generalization}
\resizebox{\linewidth}{!}{ 
\begin{tabular}{l c c c}
\toprule
\textbf{Method} & \textbf{Backbone} & \textbf{rFID} $\downarrow$ & \textbf{gFID} $\downarrow$ \\
\midrule
\textbf{RePack} (Ours) & DiT-XL/1 (64 ep) & 0.83 $\to$ \textbf{0.69} & 1.82 $\to$ \textbf{1.65} \\
\midrule
VA-VAE~\cite{VAVAE} & DiT-XL/1 (64 ep) & 0.28 $\to$ \textbf{0.22} & \cellcolor{red!10} 2.11 $\to$ 2.86 $\uparrow$ \\
SD-VAE~\cite{DiT} & DiT-XL/2 & 0.60 $\to$ \textbf{0.40} & \cellcolor{red!10} 2.27 $\to$ 2.51 $\uparrow$ \\
\bottomrule
\end{tabular}
}
\vspace{-10pt} 
\end{wraptable}

As shown in Table~\ref{tab:refiner_generalization}, While the Refiner consistently lowers the reconstruction error (rFID) for all methods, it degrades the generative performance (gFID) for both VA-VAE ($2.11 \to 2.86$) and SD-VAE ($2.27 \to 2.51$).
This contrasts with RePack, where the Refiner yields a significant boost ($1.82 \to 1.65$).

\paragraph{Hypothesis: Distribution Mismatch.}
We hypothesize that the performance drop mainly comes from a mismatch between the Refiner’s training and inference conditions.
The Refiner is trained using clean, ground-truth latents and their decoded images, whereas at inference, it must operate on latents generated by the DiT, which are noisier and follow a different distribution.
For SD-VAE and VA-VAE, this gap is especially large: their ground-truth latents are detailed (rFID $\approx$ 0.2–0.6), which likely leads the Refiner to depend on precise cues that are no longer reliable once the latent is generated (gFID $>2.1$).

In contrast, RePack builds on VFM features that naturally emphasize high-level semantics rather than low-level textures.
The compact bottleneck ($d=32$) further suppresses fine-grained information, intentionally producing a latent representation that retains semantic structure while discarding detailed appearance.
This representation matches the role of the Refiner, which is responsible for reconstructing high-frequency details from coarse semantic cues.
As a result, unlike other VAEs where semantics and texture are entangled, RePack enforces a clear separation of responsibilities, allowing the Refiner to function as a necessary and well-aligned component.
Nevertheless, designing a universal Refiner is a promising direction for future work, which we plan to further explore.

\subsection{Efficiency under Resource Constraints.}
It is worth noting that our current experiments are constrained by limited computational resources, restricting our training to a maximum of 80 epochs.
In contrast, most state-of-the-art baselines (e.g., RAE~\cite{RAE}, SiT~\cite{SiT}, MAR~\cite{MAR}) rely on prolonged training schedules ranging from 800 to 1,600 epochs to fully converge.
Despite this significant disparity (training for only $\sim$5-10\% of the standard schedule), RePack equipped with the Refiner achieves a highly competitive FID of 1.61.
As shown in Table~\ref{tab:repack_full_exp_results}, this performance is comparable to several competitive methods trained for hundreds of epochs.
We acknowledge this as a current limitation and plan to extend the training in future work.

Furthermore, these constraints on compute and time also limited our scope regarding Text-to-Image (T2I) generation.
We acknowledge this limitation.
Extending RePack to large-scale T2I generation remains a primary direction for our future research.

\section{Extended Discussion on Concurrent Approaches}
\label{sec:extended_discussion}

In this section, we provide a comparison with concurrent works that similarly leverage Vision Foundation Models (VFMs) for generation.

Integrating Vision Foundation Models (VFMs) into generative pipelines has recently attracted increasing attention.
Concurrent with our work, several studies \cite{VFM-VAE, PS-VAE, FAE} independently explore similar directions.
The emergence of these efforts reflects a shared view in the community that the semantic representations learned by VFMs can play an important role in improving the efficiency of generative models.

While sharing this broader motivation, these concurrent works diverge in their specific methodologies and trade-offs.

\textbf{VFM-VAE} \cite{VFM-VAE} focuses on architectural modifications to the decoder to handle coarse features.
While effective, it requires more training to align the representation, achieving an FID of 1.62 after 640 epochs.
In comparison, RePack achieves a comparable FID of 1.61 (with Refiner) in 80 epochs.

\textbf{FAE} \cite{FAE} introduces attention-based adapters to bridge the feature gap, leveraging massive backbones like DINOv2-Giant to ensure performance.
RePack achieves competitive results (FID 1.82 vs. 1.87 at 64 epochs) using a standard ViT-B backbone and a lightweight linear projector, offering a more parameter-efficient solution.

\textbf{PS-VAE} \cite{PS-VAE} adopts a joint fine-tuning strategy and compresses features to 96 channels.
Although direct generation comparison is unavailable due to differing benchmarks, our empirical analysis suggests that a compact 32-channel manifold (vs. 96) is more beneficial to rapid DiT convergence.
Furthermore, it remains unclear whether fine-tuning the VFM consistently preserves the generality of VFM representations, as it may lead to dataset-specific adaptation.

Ultimately, we acknowledge that each of these concurrent approaches offers unique insights and advantages to the field.
We hope that RePack, alongside these pioneering works, contributes to a broader research direction that pushes the boundaries of efficient, VFM-driven generative models.

\section{Further Clarifications}
\label{app:clarifications}
In this section, we provide clarifications on the conceptual visualization, feature dimensionality, reconstruction versus generation quality, and reproducibility of RePack.

\subsection{Clarification on Conceptual Visualization}
\label{app:visualization_note}
We clarify that the schematic illustration of the optimization landscape (Figure~\ref{fig:design_concept}, bottom) is AI-generated and is intended solely to convey our design philosophy. 
The actual feature spaces operate in much higher dimensions. 
The raw VFM space (DINOv3-B) contains 768 channels, and the compressed RePack manifold still consists of 32 channels. 
Direct visualization of such high-dimensional spaces is difficult. 
Therefore, we adopt a dimensional analogy, the rugged 3D terrain represents the complexity of the high-dimensional space, and the glowing planar slice simulates the projection onto a lower-dimensional manifold. 
Ultimately, this illustration is meant to convey that, compared to the more simpler optimization path in the compact RePack space, fitting a distribution in the large and redundant raw VFM space requires a larger and complex search space.

\subsection{Clarification on Feature Dimensionality and Redundancy}
In the Introduction Section (Main Text), we note that standard VFM features (e.g., DINOv3 with shape $768 \times H/16 \times W/16$) contain the same number of numerical elements as the original RGB image ($3 \times H \times W$). This comparison is intended to highlight the \textbf{dimensional redundancy} of the raw VFM latent space, rather than to equate the information content of VFM features with raw pixels.
While VFM features are semantically structured and low-rank, standard VFM encoders embed them into a high-dimensional space (e.g., 768 channels), resulting in a clear dimensional redundancy.

\subsection{Clarification on Reconstruction vs. Generation Quality}
\label{app:reconstruction_discussion}
We frankly acknowledge that, compared to VAEs trained end-to-end in pixel space, using frozen VFM features leads to weaker reconstruction fidelity. However, prior work such as SVG~\cite{SVG} shows that pursuing better reconstruction (e.g., reducing rFID from 1.17 to 0.65) may yield only marginal improvements in generation quality (gFID 6.12 → 6.11; Table 4 in the SVG paper). 

RePack emphasizes semantic packing at the encoding stage, while assigning high-frequency detail recovery to the Refiner. This design is motivated by our goal of enabling efficient generative modeling while maintaining high image fidelity. As a result, this decoupled strategy improves both reconstruction performance (rFID from 0.83 to 0.69) and generation quality (gFID from 1.82 to 1.65).

Nevertheless, we also acknowledge that this strategy may introduce limitations when applied to downstream tasks such as image editing. In future work, we plan to explore more flexible spatial control mechanisms within this framework to maintain generation efficiency while extending its applicability to fine-grained editing and other downstream tasks.

\subsection{Reproducibility and Code Availability}
We emphasize the importance of reproducibility and open-source community engagement.
To ensure that the community can fully reproduce our results, we provide our complete implementation, including the source code, training configurations, and pretrained model checkpoints.
The repository is available at \url{https://github.com/guanfangdong/RePack-then-Refine}.

\section{Additional Visualizations}
\label{sec:supp_vis}

\subsection{Extended Generated Samples}
\begin{figure*}[h]
    \centering
    \includegraphics[width=0.85\textwidth]{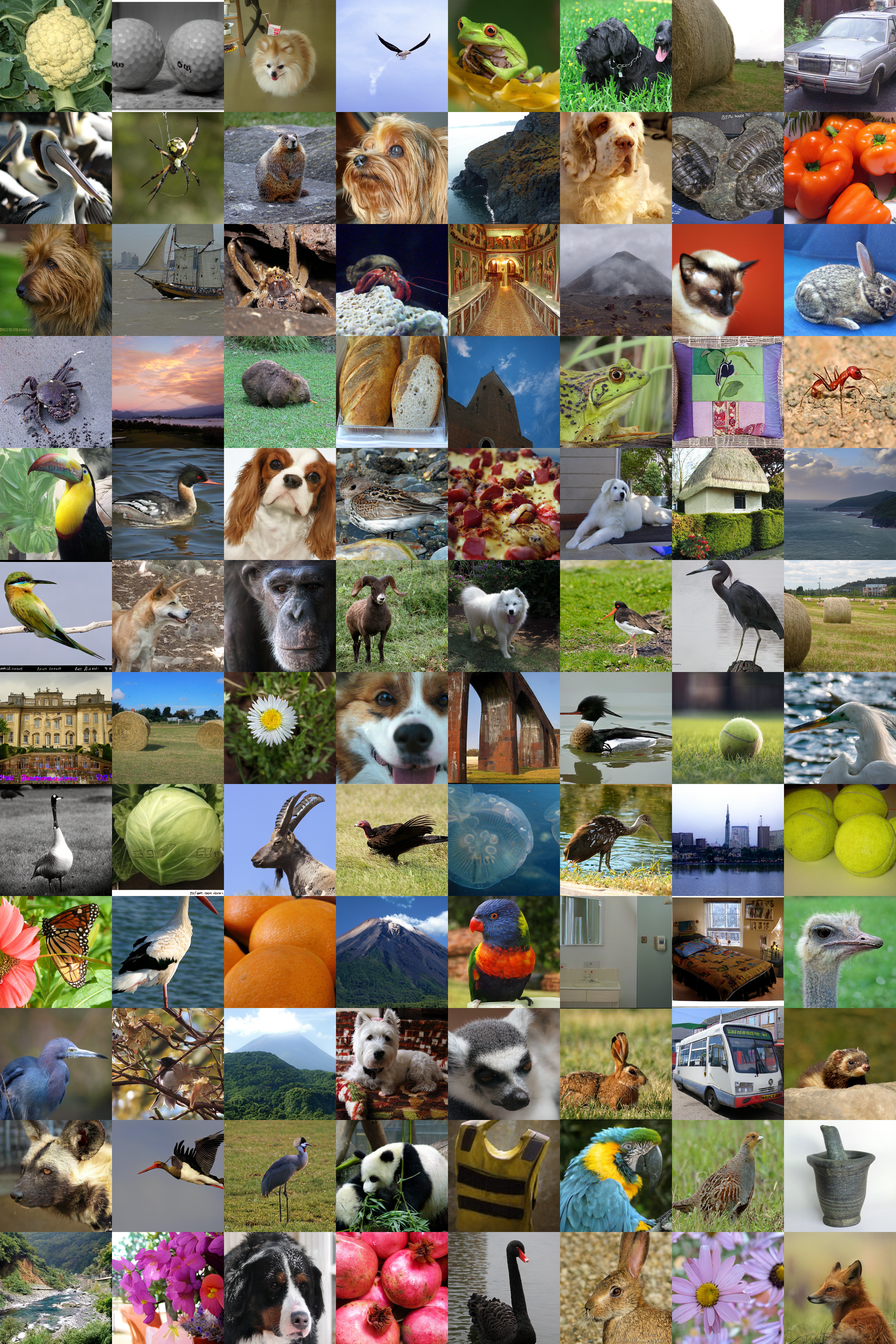}
    \caption{\textbf{Extended Generated Samples (64 Epochs).} Samples generated by RePack-DiT-XL/1 + Refiner on ImageNet $256 \times 256$.}
    \label{fig:large_grid}
\end{figure*}

Figure \ref{fig:large_grid} presents samples generated by RePack-DiT-XL/1 with the Latent-Guided Refiner after 64 training epochs. 

\subsection{Training Progression}
\begin{figure*}[h]
    \centering
    \includegraphics[width=0.83\textwidth]{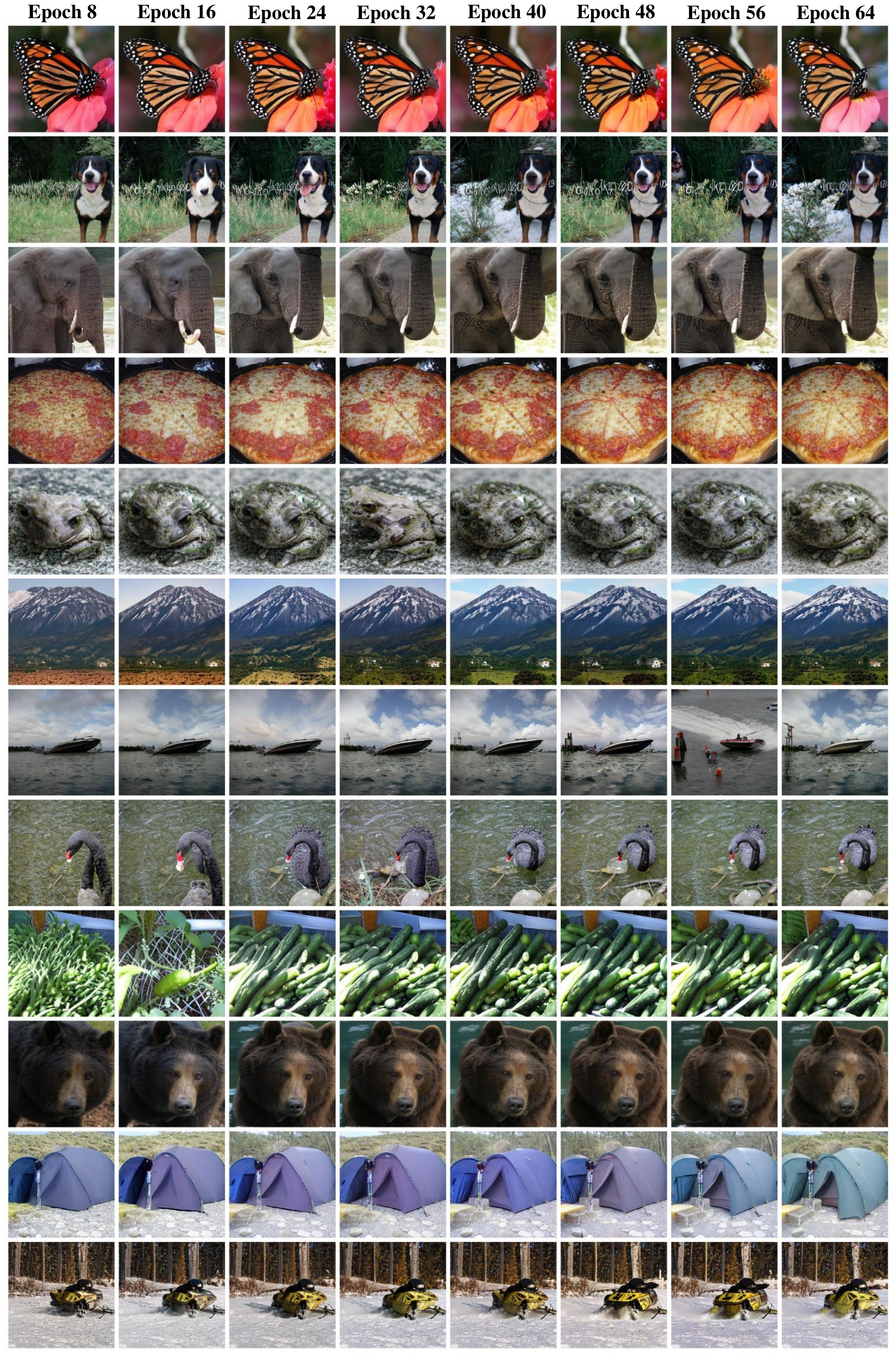}
    \caption{\textbf{Visualization of Training Progression.} Samples generated by RePack-DiT-XL/1 at 8, 16, 24, 32, 40, 48, 56 and 64 epochs. The model exhibits rapid convergence, forming stable semantic structures at a very early stage.}
    \label{fig:training_process}
\end{figure*}

Figure \ref{fig:training_process} visualizes the generation quality of RePack-DiT-XL/1 across different training stages, covering from 8 to 64 epochs.
Notably, the model establishes a coherent semantic structure as early as epoch 8, producing recognizable object shapes and layouts.

\subsection{Extended Analysis of Latent Representations}
\begin{figure*}[t]
    \centering
    \includegraphics[width=0.85\textwidth]{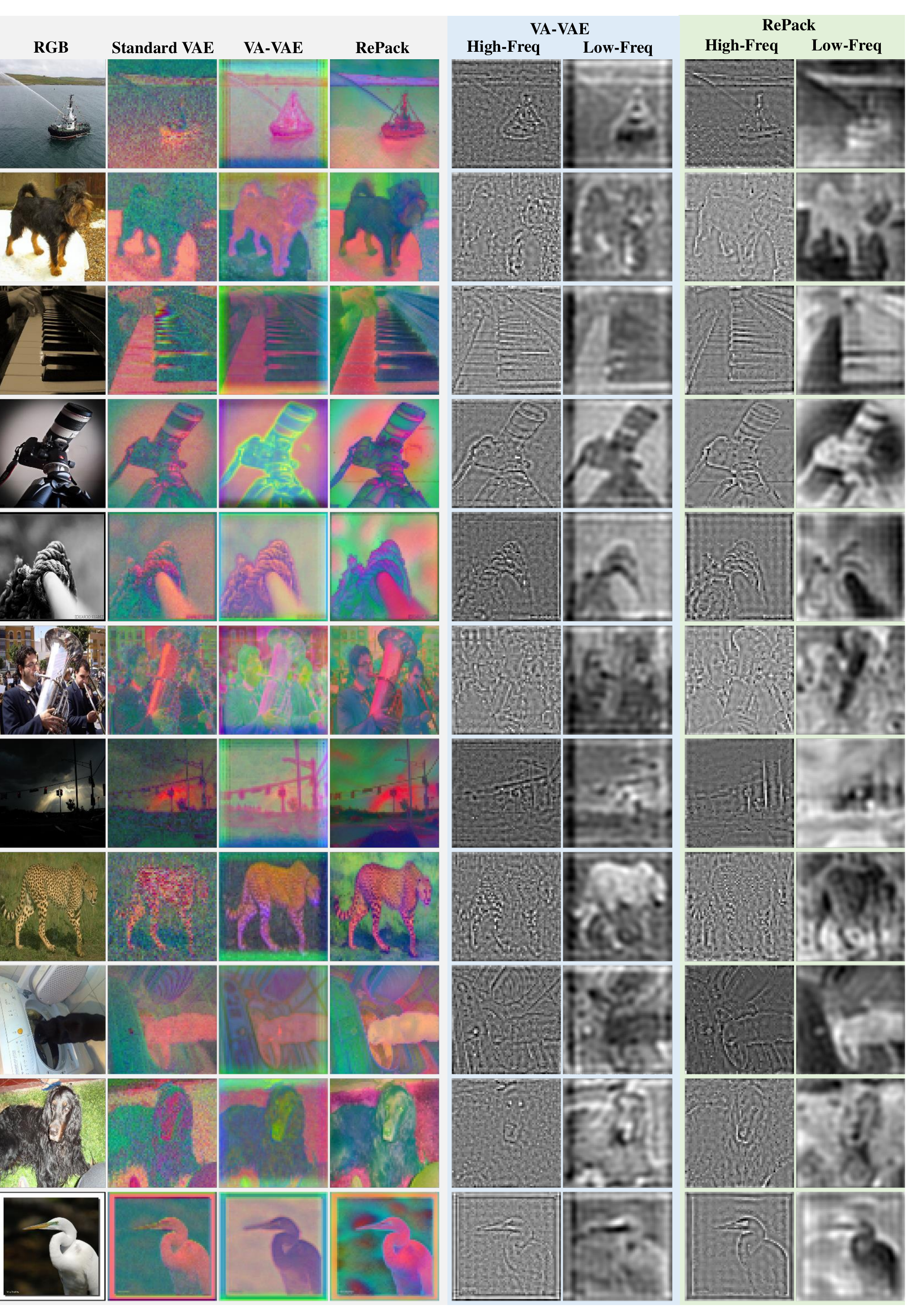}
    \caption{\textbf{Extended Visual Analysis of Latent Representations.} 
    Complementing Figure \ref{fig:latent_vis} in the main text, this figure provides additional visualizations.
    \textbf{Left:} Input images. 
    \textbf{Middle:} PCA visualizations of latents from a Standard VAE (trained w/o VFM guidance), VA-VAE, and RePack. 
    \textbf{Right:} Spectral decomposition of VA-VAE and RePack latents. }
    \label{fig:latent_vis_ext}
\end{figure*}

Figure \ref{fig:latent_vis_ext} provides extended PCA visualizations and spectral decompositions, further verifying that RePack produces more semantically coherent and structurally clean latent representations.

\end{document}